%% file: main.tex
\documentclass[10pt,twocolumn,letterpaper]{article}

\usepackage{cvpr}              % To produce the CAMERA-READY version

\definecolor{cvprblue}{rgb}{0.21,0.49,0.74}
\usepackage[pagebackref,breaklinks,colorlinks,allcolors=cvprblue]{hyperref}
\usepackage{algorithm}
\usepackage{multirow}
\usepackage{algpseudocode}
\usepackage{booktabs}   
\usepackage{amsmath}
\def\paperID{2651} % *** Enter the Paper ID here
\def\confName{CVPR}
\def\confYear{2025}

\title{VISTA3D: A Unified Segmentation Foundation Model For 3D Medical Imaging}
\author{Yufan He$^{1}$ , Pengfei Guo$^{1}$ , Yucheng Tang$^{1}$ , Andriy Myronenko$^{1}$ , Vishwesh Nath$^{1}$,\\ Ziyue Xu$^{1}$, Dong Yang$^{1}$ , Can Zhao$^{1}$ , Benjamin Simon$^{3,4}$ , Mason Belue$^{2}$, \\Stephanie Harmon$^{3}$ , Baris Turkbey$^{3}$ , Daguang Xu$^{1}$ , Wenqi Li$^{1}$  \\
$^{1}$ NVIDIA \\ $^{2}$ University of Arkansas for Medical Sciences \\ $^{3}$National Institutes of Health \\ $^{4}$ University of Oxford}
\begin{document}
\maketitle
\input{sec/0_abstract}    
\input{sec/1_intro}

\input{sec/2_method}

\input{sec/3_result}

\input{sec/4_discussion}

{
    \small
    \bibliographystyle{ieeenat_fullname}
    \bibliography{main}
}

\end{document}

\input{sec/X_suppl}    
{
    \small
    \bibliographystyle{ieeenat_fullname}
    \bibliography{main}
}


%% file: sec/0_abstract.tex
\begin{abstract}
    Foundation models for interactive segmentation in 2D natural images and videos have sparked significant interest in building 3D foundation models for medical imaging. However, the domain gaps and clinical use cases for 3D medical imaging require a dedicated model that diverges from existing 2D solutions. Specifically, such foundation models should support a full workflow that can actually reduce human effort. Treating 3D medical images as sequences of 2D slices and reusing interactive 2D foundation models seems straightforward, but 2D annotation is too time-consuming for 3D tasks. Moreover, for large cohort analysis, it's the highly accurate \textbf{automatic} segmentation models that reduce the most human effort. However, these models lack support for interactive corrections and lack zero-shot ability for novel structures, which is a key feature of ``foundation". While reusing pre-trained 2D backbones in 3D enhances zero-shot potential, their performance on complex 3D structures still lags behind leading 3D models. To address these issues, we present VISTA3D, \textbf{V}ersatile \textbf{I}maging \textbf{S}egmen\textbf{T}ation and \textbf{A}nnotation model, that targets to solve all these
    challenges and requirements with one unified foundation model. VISTA3D is built on top of the well-established 3D segmentation pipeline, and it is the first model to achieve state-of-the-art performance in both 3D automatic (supporting 127 classes) and 3D interactive segmentation, even when compared with top 3D expert models on large and diverse benchmarks. Additionally, VISTA3D's 3D interactive design allows efficient human correction, and a novel 3D supervoxel method that distills 2D pretrained backbones grants VISTA3D top 3D zero-shot performance. We believe the model, recipe, and insights represent a promising step towards a clinically useful 3D foundation model. Code and weights are publicly available at \url{https://github.com/Project-MONAI/VISTA}.
\end{abstract}

%% file: sec/1_intro.tex
\section{Introduction}
Three-dimensional medical imaging such as computed tomography (CT) is widely used for creating cross-sectional volumetric images within various body regions. As a major anatomic imaging modality, it reveals detailed morphological information of body structures and abnormalities. In clinical practice, manual segmentation is time-consuming and tedious, thus developing better automatic models has been one of the most active research topics. A typical direction is enhancing network architecture and tailoring training recipes for specific tasks~\cite{myronenko2018brats,he2021dints,tang2022self,nnUNet}. For each task, curating a specific set of training data and training expert models is often performed, which requires strong engineering expertise. A model that solves a variety of tasks out of the box is thus more desirable.

\begin{figure*}
    \centering
    \includegraphics[width=1\textwidth]{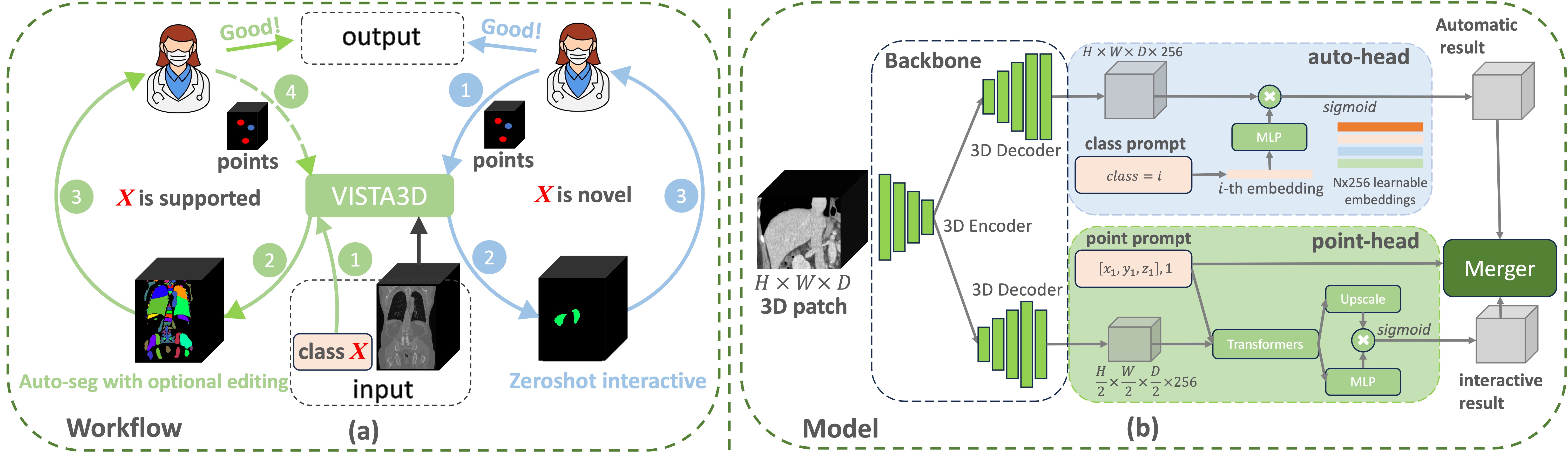}
    \caption{Fig.(a) shows the full human-in-the-loop workflow VISTA3D supports. If the segmentation task $X$ is within 127 supported classes~(left green circle), VISTA3D performs accurate automatic segmentation. The doctor can inspect and efficiently edit the result  with VISTA3D if needed. If $X$ is a novel class~(right blue circle), VISTA3D  performs 3D interactive zero-shot segmentation. Fig.(b) shows the VISTA3D architecture. It contains two branches that share the same image encoder. The top auto-branch will activate out-of-the-box automatic segmentation if user provide a class prompt that's within 127 supported classes. The bottom interactive branch will activate interactive segmentation if user provide 3D point click prompts. If both branches are activated, a merger module based on Alg.~\ref{alg:ir} will use interactive results to edit automatic results.}
    \label{fig:whole-flow}
\end{figure*}
Unlike natural images where there could be an unlimited number of object classes, the clinically relevant healthy human anatomies revealed by CT or MRI are limited (such as liver, pancreas), thus training an \textbf{automated} segmentation model that supports most of standard human anatomies is technically feasible~\cite{TotalSegmentator,liu2023clip,ji2023continual,zhao2023one,ulrich2024multi}. However, in practice, clinicians may be more interested in rare pathologies or animal data that are usually unsupported by those models due to data scarcity. Lacking zero-shot capability to handle those use cases becomes a significant limitation. Meanwhile, it is important for the model to allow human input for correction for procedures like surgical planning.

Recently, large language models~\cite{team2023gemini,achiam2023gpt,touvron2023llama} have shown strong generalizability on various tasks and are considered the foundation models. The idea of a ``promptable'' system has been proposed to achieve a flexible model that can solve different tasks out-of-the-box. For image segmentation, Segment Anything~(SAM)~\cite{kirillov2023segment} has gained great interest and achieved impressive zero-shot performance. In the medical domain, recent work~\cite{MedSAM} hence adapted SAM to medical imaging modalities via model fine-tuning. These SAM-based methods demonstrate promising results in 2D that leverage interactive user input. For 3D medical images, such prompt (e.g. point) binding to every class, every slice, and every scan, often requires substantial human effort, making it infeasible for large cohort data analysis. The recent Segment Anything in Video~(SAM2)~\cite{ravi2024sam} triggered even greater interest since a 3D scan is represented by a stack of 2D cross-sectional images (slices), while a video is also a stack of 2D images (frames). However, our experiments show that the SAM2 framework, even well-finetuned on 3D medical datasets, cannot compare with VISTA3D, especially on complicated 3D structures~(details in the supplementary).  SAM2 tracks objects over time, but medical imaging requires spatially consistent treatments for volumetric inputs. For example, a car in different time frames is still the same car, but its internal 2D cross-sectional images can be completely different objects like seats and engines. This illustrates the big gap between 2D natural images or videos and cross-sectional medical images. Similarly, SAM3D~\cite{bui2024sam3d} extracts 3D volume features slice by slice with a 2D SAM encoder and a 3D decoder, but the results are much worse than 3D experts. Simply applying methods from natural images to 3D medical images will fall short.

Recent works exploring in-context learning for medical image segmentation~\cite{shen2024segicl,Wu_2024_CVPR,butoi2023universeg,ren2024medical} can segment any class guided by example image or text. This seems like an optimal case because it does not require model finetuning or time-consuming human input. However, the performance of such methods is far behind~\cite{ren2024medical} the dataset-specific supervised models (e.g. nnU-Net ~\cite{nnUNet}).  

We envision that a foundation model for 3D medical image segmentation should support a full workflow (Fig.~\ref{fig:whole-flow}(a)) that can reduce human effort, which may have the following essential capabilities: 1) Highly accurate automatic segmentation for common organs or structures; 2) Ability to interact with human experts, allowing for effective refinements of existing segmentation results; 3) Zero-shot capabilities, either allow the user to interactively annotate unseen classes or use in-context learning via text or example guidance. The model should operate in 3D since 2D slice-by-slice methods are too time-consuming and may not fully leverage 3D visual contexts; 4) Few shot/transfer learning abilities that allow users to quickly finetune the model to perform accurate automatic segmentation on new classes, given that existing in-context learning or open vocabulary segmentation still fall short compared to expert 3D models in accuracy.

To support this workflow and achieve comparable performances with the best expert models, we should build the model based on well-established 3D pipelines that rely on 3D backbones and sliding window inference. However, this direction is not taking advantage of the existing 2D pretrained weights with a strong zero-shot ability(e.g. SAM). Reusing SAM weights and adding light-weight 3D adaptor modules~\cite{wu2023medical,gong20233dsam} seems viable but the automatic performance on diverse classes~(comparing to TotalSegmentator~\cite{TotalSegmentator}) are limited due to freezing the majority of weights. So the challenge is, how to build a model that possesses the advantage of well-established 3D pipelines, while also utilizing the insights and checkpoints from 2D natural images to solve 3D problems. Based on this goal, we introduce VISTA3D, and our contributions are:  
\begin{enumerate}
    \item The first unified foundation model that supports a full annotation workflow, and achieves state-of-the-art 3D promptable automatic segmentation and interactive editing, benchmarked over 14 challenging datasets with 127 classes and compared with well-established baselines.
    \item A novel supervoxel methods are developed to distill 2D foundation models for 3D medical imaging, which boosted VISTA3D's zero-shot performance by 50\% and achieved state-of-the-art 3D zero-shot performance with much less annotation efforts.
    \item We curated a large CT dataset with 11454 scans, paired it with partial manual labels, pseudo labels, and supervoxels, and proposed a novel four-stage training recipe to tackle the challenges to achieve state-of-the-art performances and editing experiences.
\end{enumerate}

% Much less annotation effort needed for major human body parts (supported classes):
% Support automatic segmentation with better accuracies/generalizability with large class numbers.
% Support interactive mask editing to correct auto results
% Support high-accuracy/generalizability single-click point annotation.
% Much less annotation effort for zero-shot classes.
\section{Related Work}
\noindent\textbf{Dataset-specific supervised training.} %nnunet, auto3dseg
Many existing 3D medical imaging segmentation methods~\cite{myronenko2018brats,he2021dints,tang2022self,nnUNet} are proposed to train dedicated models for a specific dataset. nnU-Net~\cite{nnUNet} is a well-established framework that can automatically adapt to different datasets. This adaptation occurs seamlessly, mitigating the need for manual intervention or specialized expertise, which expedites the medical imaging segmentation models.  Auto3DSeg\footnote{\url{https://monai.io/apps/auto3dseg}} presents a holistic approach to tackling the challenge of large-scale 3D medical image segmentation. It also provides automatic task adaptation. These two frameworks have proven their effectiveness by winning numerous highly competitive 3D segmentation challenges~\cite{nnUNet, gatidis2023autopet,dorent2024lnq,payette2024multi,myronenko2023aorta,myronenko23brats23,myronenko2023kits23,myronenko23sega,myronenko2023automated}. Although auto-configuration solutions can speed up the curation of task-specific expert models and achieve high performance, they lack inherent zero-shot capabilities and require human effort and resource in data preparation and training.

\noindent\textbf{Foundational segmentation model.} 
%Totalsegmentator, TotalSegmentator: Robust Segmentation of 104 AnatomicalStructures in CT images
% Universal model CLIP-Driven Universal Model for Organ Segmentation and Tumor Detection
% Continual Segment: Towards a Single, Unified and Non-forgetting Continual Segmentation Model of 143 Whole-body Organs in CT Scans, e.t.c. 
Foundation segmentation models aim to develop a single unified deep learning model capable of segmenting multiple anatomical structures/organs from whole-body CT scans, rather than training separate models for each organ. Totalsegmentator~\cite{TotalSegmentator} is proposed for fully automatic segmentation of over 117 anatomical structures in CT images covering various organs, bones, muscles, and vessels. It represents a significant contribution to the biomedical imaging community, enabling researchers and clinicians to leverage accurate and comprehensive segmentation without requiring time-consuming manual efforts. The Universal Model~\cite{liu2023clip} leverages text embeddings from the CLIP model to encode the anatomical relationships between organs and tumors and support the automatic segmentation of 31 classes. SAT~\cite{zhao2023one} supports automatic segmentation on 497 classes based on text prompts. Although they tried to incorporate text embeddings into automatic segmentation, the text prompts only work with a fixed vocabulary and do not support zero-shot or open-vocabulary segmentation. Continual Segment~\cite{ji2023continual} is a unified model capable of segmenting 143 body organs by a frozen encoder coupled with incrementally added decoders to avoid catastrophically forgetting previously learned structures. While those foundational 3D segmentation models represent significant advancements in multi-organ segmentation, the inability of zero-shot and interactive segmentation impedes their real-world applicability. The in-context learning or open vocabulary segmentation ~\cite{shen2024segicl,Wu_2024_CVPR,butoi2023universeg,ren2024medical} are desirable, which can achieve automatic segmentation on the unseen class by prompts or support examples. However, at the current stage, their performances fall short compared to the expert models, especially for 3D images~\cite{ren2024medical}.

\noindent\textbf{Interactive medical image segmentation.} %MedSAM~\cite{} finetuned the SAM model on a diverse 2D medical images, which supports 2D bounding box annotation. segvol, SAM-MED3D, e.t.c 
The Segment Anything Model (SAM)~\cite{kirillov2023segment} and its video variant~(SAM2)~\cite{ravi2024sam} have inspired and enabled various medical imaging applications through the adaptation and fine-tuning of medical data~\cite{MedSAM,gong20233dsam,wu2023medical,huang2024segment}. MedSAM~\cite{MedSAM} finetunes SAM on large 2D medical datasets using the bounding box but lacks the ability for detailed editing and handling 3D inputs. The SAM adapters~\cite{gong20233dsam,wu2023medical} add 3D adaptor modules to the SAM backbone for efficient 3D finetuning, however, the 3D performance was validated on limited classes and benchmarks. The work closely related to ours is SegVol~\cite{du2023segvol}, a 3D foundation model designed for 3D semantic and interactive segmentation. However, SegVol~\cite{du2023segvol}'s performance relies heavily on the 3D bounding box added with text prompts, and there is still a big performance gap with its text prompt-based automatic segmentation. Although those work approved their effectiveness in segmenting 10 to 20 classes of 3D structures on benchmarks like BTCV~\cite{landman2015miccai} or AMOS~\cite{ji2022amos}, the problem is, that those structures can be easily solved by automatic foundation models like TotalSegmentator~\cite{TotalSegmentator}. It is important to rethink the position of interactive models and how they can really reduce human efforts for 3D medical segmentation.

%% file: sec/2_method.tex
\section{Method} 

\subsection{Overview}
 % \begin{figure}
 %     \centering
 %     \includegraphics[width=0.37\textwidth]{F/diagram.png}
 %     \caption{The overall workflow VISTA3D supports.}
 %     \label{fig:diagram}
 % \end{figure}
We separate the segmentation tasks into \textbf{supported classes} and \textbf{zero-shot classes}. The supported classes are the classes that have enough training data with annotations with which we can train VISTA3D to perform automatic segmentation (here we support 127 classes). We curated a global class index list and mapped the groundtruth indices from those partially annotated datasets to this list. We trained the automatic head to accept the index as the prompt and output a binary segmentation. For zero-shot classes, the segmentation is mainly generated by the VISTA3D interactive branch, which accepts user click coordinates in 3D. The interactive segmentation also works for supported classes. The overall workflow is shown in Fig.~\ref{fig:whole-flow}(a). To train such a model, we curated a large dataset containing 11454 3D CT scans, generated pseudo labels from TotalSegmentator model~\cite{TotalSegmentator} and supervoxels using SAM pre-trained weights~\cite{kirillov2023segment}(see detail in Sec \ref{sec:data}). A stage-by-stage training recipe is used to train interactive and automatic workflows systematically. 

 % \begin{figure}
 %     \centering
 %     \includegraphics[width=0.49\textwidth]{F/framework.png}
 %     \caption{The VISTA3D model contains two branches that share the same image encoder. The top auto-branch performs out-of-the-box automatic segmentation for 127 supported classes. The bottom interactive branch accepts user clicks and performs interactive segmentation on both supported classes and novel zero-shot classes.}
 %     \label{fig:model}
 % \end{figure}
 
\subsection{Model architecture}

SAM's~\cite{kirillov2023segment} image encoder is a vision transformer (ViT)~\cite{dosovitskiy2021vit} with 16$\times$16 patch embedding. For 3D images, ViT becomes extremely memory-demanding, since the token length (number of patches) is much longer compared to 2D images. On the other hand, 16$\times$16 patch embedding inevitably loses spatial details.  Computationally feasible adaptations of transformers to 3D images have been proposed~\cite{hatamizadeh2022unetr,yufanhe2023swingunetr2}, but the state-of-the-art results, as shown in the recent MICCAI 3D segmentation challenges, are still predominantly based on the convolutional architectures. Specifically, the SegResNet model, a U-net type architecture, has won BraTS 2023~\cite{myronenko2018brats}, KiTS 2023~\cite{myronenko2023kits23} and Seg.A 2023~\cite{myronenko23sega} MICCAI 3D segmentation challenges. In VISTA3D, we use SegResNet~\cite{myronenko2018brats} from MONAI~\cite{cardoso2022monai}, as a backbone CNN, and followed the best practices in medical image segmentation of patch-based training~(we used 128-voxel cubic patch) and sliding window inference. 

\noindent\textbf{Automatic branch} As shown in Fig.~\ref{fig:whole-flow}(b), there are two branches: the automatic branch (top) and the interactive branch (bottom). The SegResNet encoder is shared between two branches for learning image embedding. Each branch has its own decoder with a skip connection with the shared encoder. The auto-head contains an MLP layer $M$ and a learnable $N\times C$ class embedding $E_c$~(we use $C=256$), where $N$ represents $N$ supported classes. The output feature $F$ from the decoder is of size $C\times H \times W \times D$. If the user wants to segment class $i$ (this single number $i$ is the input prompt for the auto-head), the corresponding class embedding $E_c[i]$ is used to map the feature into segmentation logits, $ sigmoid(M(E_c[i]) \times F)$. Compared to models that output all classes and apply $softmax$ at the end, our promptable scheme reduces memory usage dramatically if number of classes are huge. Meanwhile, it avoids the partial label problem in training on diverse partially labeled datasets. We added this additional MLP layer $M$ due to empirically better performance. 

Works~\cite{du2023segvol,liu2023clip,zhao2023one} have tried to use text embedding like CLIP but none of them are able to achieve zero-shot or open vocabulary segmentation with text, which is the main benefit of text prompts. Moreover, we empirically found CLIP-embedding gives slightly worse results than randomly initialized class embeddings~(used by VISTA3D).

\noindent\textbf{Interactive branch}  For the interactive branch, the click points' 3D coordinates and their labels (positive or negative) are accepted as prompts for the point head. The point head is based on the SAM's~\cite{kirillov2023segment} point prompt encoder, where the feature map performs cross-attention with point embedding. We made several changes to satisfy the needs for 3D medical images: 1) To keep the high-resolution details, the point head input feature is first upsampled back to the original image resolution with the long skip connection and then 2x downsampled to reduce the memory footprint. All the related operations including point embedding are changed to 3D. The downsampled feature and the point embedding will go through cross-attention transformers and generate the final output. 2)
To increase the click response speed for a better user experience, only a local patch centered at the click point will be segmented and used to refine automatic results. 3) For some classes that have ambiguity or overlap, e.g. pancreas/pancreas tumor and colon/colon tumor, a single point click cannot solve the ambiguity. Since the model knows the class $x$ to segment beforehand, we can add a special embedding to the click point automatically if $x$ lies in specific classes like colon/pancreas tumor. This embedding can be used to distinguish ambiguous classes. Note that this special embedding is the same for all classes with ambiguity. If we use class embedding in $E_c$ to solve the ambiguity, the point head will learn a shortcut to ignore point clicks. 4) Another challenge is that the interactive branch needs to handle both supported classes and unseen classes (zero-shot), while there might be a conflict between these two tasks. Segmenting supported class with high accuracy will require the model to remember or overfit specific features about the class like the shape and position. However, organ-specific tuning could hurt zero-shot generalizability. We mitigate this problem by adding a zero-shot embedding to the point head cross-attention if the class $x$ is a novel class. Examples can be found in the supplementary.

\noindent\textbf{Interactive refinement over automatic results} As can be seen in Fig.~\ref{fig:whole-flow}(b), the two branch outputs are independent from each other. The use case by combining the results is to interactively correct the automatic segmentation results. As illustrated in FocalClick~\cite{chen2022focalclick}, the interactive refinement over existing masks could destroy the correct part. We observe this behavior when simply combining the interactive results and the automatic results. We used the local refinement idea from FocalClick and proposed the following merging algorithm ~Alg.~\ref{alg:ir}. The core idea is to add or remove only the connected component regions that contain the point clicks to avoid unexpected modification. 

% \noindent\textbf{Model design choice}
% The most straightforward way to allow both automatic and interactive segmentation is to use the same backbone, same prompt encoder but different prompt types, which can be class-prompt~(class-specific embedding like one-hot embedding or CLIP embedding~\cite{liu2023clip}) and interactive prompt~(e.g. points, boxes). But there are intrinsic conflicts 

\begin{algorithm}
\caption{Interactive refinement on the automatic results}\label{alg:ir}
\begin{algorithmic}
\Require $I$ positive and $J$ negative clicks $P^i_p$ and $P^j_n$, automatic and interactive output $M_a$ and $M_p$
\Ensure $size(M_a) = size(M_p)$
\State Denote ``get 3D connected components" as $CC(\cdot)$.
\State $\{M^n_{add} \}_N \gets CC((M_p - M_a) > 0) $ \Comment{N added connected components}
\State $\{M^k_{rm} \}_K \gets CC((M_a - M_p) > 0) $ \Comment{K removed connected components}
\If{$\underset{i={1,\cdots,I}}{\exists}{P^i_p} \in M^a$} $M^n_{add} = M^n_{add} \cup CC(M^p)$  \EndIf \Comment{If positive points in $M_a$, add $M_p$ into addition candidates.}
\State $ M_{final\_rm}, M_{final\_add} \gets \{\}, \{\}$
\For{$n=1$ to $N$}
\If{$\underset{i={1,\cdots,I}}{\exists}{P^i_p} \in M^n_{add}$} $M_{final\_add} = M_{final\_add} \cup M^n_{add}$  \EndIf
\EndFor
\For{$k=1$ to $K$}
\If{$\underset{j={1,\cdots,J}}{\exists}{P^j_n} \in M^k_{rm}$} $M_{final\_rm} = M_{final\_rm} \cup M^k_rm$  \EndIf
\EndFor
\State \Return $M_a + M_{final\_add} - M_{final\_rm}$
\end{algorithmic}
\end{algorithm}

\subsection{Data}\label{sec:data}
We curated a collection of 11454 CT volumetric images obtained from in-house and publicly available data sources~\cite{roth9data,roth2015deeporgan,ma2021abdomenct,ma2023unleashing,ji2022amos,antonelli2022medical,rister2020ct,TotalSegmentator,simpson2024preoperative,heller2019data,sekuboyina2021verse,stoverud2023aeropath,armato2015data,harmon2020artificial,johnson2008accuracy,saltz2021stony,national2011reduced} with a wide range of acquisition protocols and subject conditions. Among these, five of them are without labels, and the rest have various voxel-wise annotation regions of interest, including anatomical structures and lesions. We denote the ground truth from those datasets as \textbf{manual labels} or \textbf{partial labels}.  Each data source is randomly split into 64\% training, 16\% validation, and 20\% test sets. We generated \textbf{pseudo-labels} of 117 classes using TotalSegmentator~\cite{TotalSegmentator} and \textbf{supervoxels} using SAM for every scan. The unreliable pseudo-labels are removed by post-processing.
\begin{figure}
    \includegraphics[width=0.47\textwidth]{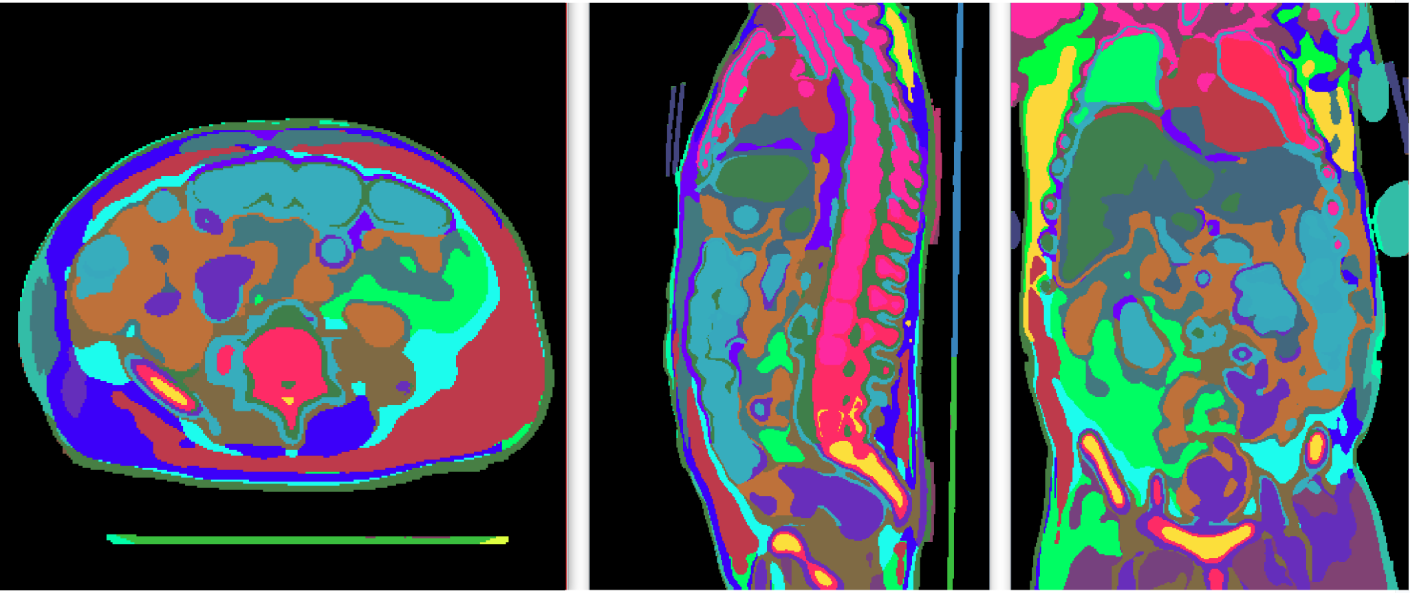} 
    \caption{Generated supervoxel from Alg.~\ref{alg:supervoxel}, showing examples in axial, sagittal, and coronal views. Different colours represent different supervoxels.}
    \label{fig:supervoxel}
\end{figure}

\noindent\textbf{Supervoxel generation}
The vast majority of SAM's zero-shot capabilities come from this large-scale supervised training on its 11 million diverse and fully annotated images~\cite{kirillov2023segment}. Those annotations helped SAM learn how humans perceive an object, and become the image segmentation foundation model. However, the manual labels or pseudo-labels in 3D CT can only cover around one or two hundred classes. We empirically found that this level of class diversity is not enough for the model to achieve SAM-like zero-shot ability in 3D. To solve this problem, most works decided to finetune SAM pretrained ViT checkpoint on 2D medical data to inherit this zero-shot ability, which inevitably limited the adaptability to 3D images. Here we propose a novel method to distill the image understanding ability from SAM by generating 3D supervoxels from 2D SAM feature maps. The algorithm is shown in Alg.~\ref{alg:supervoxel}. We perform a 3D supervoxel algorithm on the upsampled SAM feature embedding, which is generated slice-by-slice in three views. An example of generated supervoxel results is shown in Fig.~\ref{fig:supervoxel}. We generate supervoxels for all 11454 CT scans and use them to train our interactive branch, and this gives VISTA3D zero-shot capabilities.
SegVol~\cite{du2023segvol} used a similar idea but the supervoxel generation is based on graph-cut, which is still on low-level image features. Instead, VISTA3D achieved better zero-shot performance through distilling knowledge from SAM.

\begin{algorithm}
\caption{3D supervoxel generation from SAM}\label{alg:supervoxel}
\begin{algorithmic}
\Require SAM pretrained ViT-H model $\Phi$, image encoder $\Phi_E$, output scaling layer in the mask decoder $\Phi_s$.  \Comment{All SAM components related to prompts are removed}
\Ensure Input 3D CT image $V$
\State $V \gets \{x_1, x_2, \cdots, x_A\} $ \Comment{V as a stack of axial slices}
\State $V \gets \{y_1, y_2, \cdots, y_C\} $ \Comment{V  as a stack of coronal slices}
\State $V \gets \{z_1, z_2, \cdots, z_S\} $ \Comment{V  as a stack of sagittal slices}
\State $ F_A, F_C, F_S \gets \{\}, \{\}, \{\}$
\For{$i=1$ to $A$;$j=1$ to $C$;$k=1$ to $S$}
\State $F_A = F_A \cup \Phi_S(\Phi_E(x_i))$ 
\State $F_C = F_C \cup \Phi_S(\Phi_E(y_j))$
\State $F_S = F_S \cup \Phi_S(\Phi_E(z_k))$
\Comment{Generate upsampled SAM feature for each slice at each axis.}
\EndFor
\State $F_{3D} \gets F_A + F_C + F_S$ \Comment{$F_A, F_C, F_S$ are 3D tensors with the same size}
\State \Return $SLIC(F_{3D}, n_{segments}=100, sigma=3)$ \Comment{We use SLIC~\cite{achanta2012slic} algorithm from skimage}
\end{algorithmic}
\end{algorithm}

\subsection{Recipe}
The training has four stages to solve the class imbalance issues and complications between the automatic and interactive branches.

\noindent\textbf{Stage1-Interactive branch training:} 
This is the first stage of VISTA3D training, and the goal is to train a strong image encoder that can extract good and generalizable features from 3D CT images, and enable the interactive branch to have good response to point clicks.  In each iteration, the inputs contain randomly cropped 128 cubic image patches, corresponding manual labels, pseudo labels, and supervoxels. We use a point sampler~(details in supplementary) to randomly sample points and its corresponding binary segmentation mask for training. The mask is generated by combining manual labels or pseudo labels with supervoxels or supervoxels alone. The goal is to diversify the ground truth and make the model responsive to all kinds of objects and boundaries. We also followed the SAM's iterative training scheme and sampled new points from the false positive or negative regions from previous predictions to improve editing ability. We used an iteration of 5.

\noindent\textbf{Stage2-Interactive branch finetuning:} 
The data imbalance issue is severe in our curated dataset since some rare classes such as tumors are only presented in a limited number of images. Stage 1 has a large number of training iterations and if we oversample under-represented classes during stage 1, those classes will soon overfit, moreover, overfit at different iterations for different classes. We disabled oversampling in stage 1, and under-represented classes were rarely sampled~(still needs to be sampled). In stage 2, we perform a quick finetuning with specific dataset oversampling to improve low-performing classes. Meanwhile, we removed the supervoxel and unlabeled dataset in finetuning. 

\noindent\textbf{Stage3-Automatic branch training:} 
The image encoder has been trained in the previous stages with all the 3D medical annotations and SAM-generated supervoxels. The training is based on binary segmentation without any class-specific information thus we can expect the encoder to generate more generalizable features. We freeze the image encoder to avoid changes to the interactive branch and train the auto branch decoder and head. The training is a common supervised training, but we randomly sample existing class indexes from the manual labels and pseudo-labels and use the corresponding binary masks as ground truth. 

\noindent\textbf{Stage4-Automatic branch finetuning:} 
Similar to stage 2, we need to improve the performance of under-represented classes. We used MAISI~\cite{guo2024maisi} to generate synthetic data containing anomalies such as tumors and lesions to enlarge the under-represented class sample size. We sample uniformly across dataset and finetuned model for a few epochs.

%% file: sec/3_result.tex
\section{Supported classes results}
\begin{figure}[!h]
    \centering
    \includegraphics[width=0.47\textwidth]{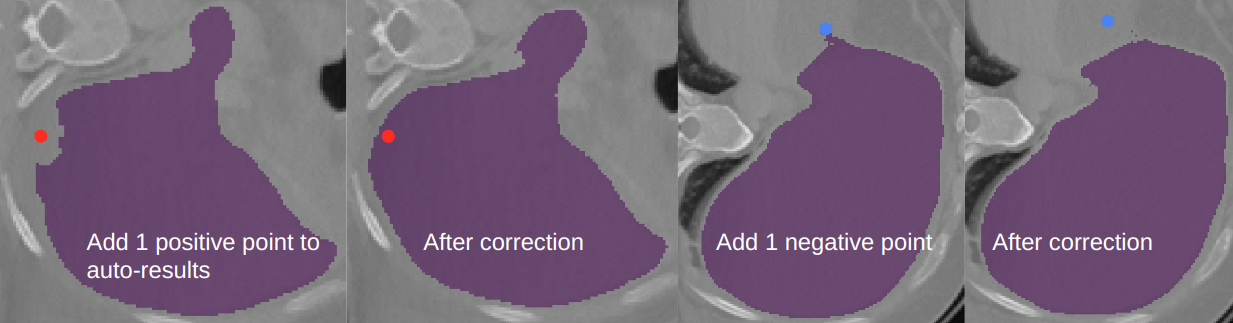}
    \caption{Correcting automatic segmentation with points. The left figure shows the automatic liver segmentation with a false negative area. After a positive point, the false negative region is corrected. The third figure shows another slice with a false positive and a negative point removed from the region shown in the last figure.}
    \label{fig:correction}
\end{figure}

\begin{figure}[!h]
    \centering
    \includegraphics[width=0.47\textwidth]{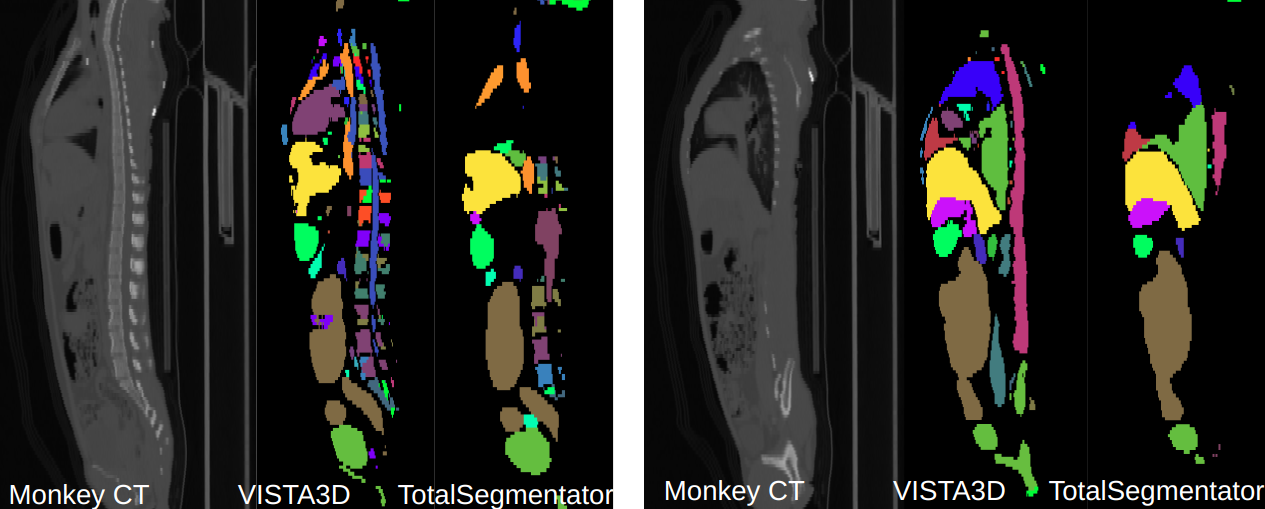}
    \caption{An example of monkey CT scan~(2 sagittal slices). We can see that VISTA3D achieved more robust segmentation.}
    \label{fig:monkey2}
\end{figure}

% \subsection{out-of-the-box automatic results}
We first test the performance of the supported classes. For supported classes, the out-of-the-box performance of a foundational model should have state-of-the-art or comparable performances to the data-specific expert models. Meanwhile, we claim VISTA3D interactive branch can correct error regions in automatic results. We show the VISTA3D's out-of-the-box automatic segmentation results~(VISTA3D auto), interactive results with a single positive click sampled from the foreground center ~(VISTA3D point), and the corrected automatic results with a single click point~(VISTA3D auto + point) with Alg.~\ref{alg:ir} in Table.~\ref{t:results}. The click point is randomly sampled from the false positive~(negative point) or false negative region~(positive point), based on which has a larger area size. Note that VISTA3D is a patch-based method using sliding window inference, thus a click point will only affect the 128 cubic patch that includes the point. The evaluation with a single point means 1 click for each sliding window patch. As for baselines, we used the Auto3Dseg framework and the nnUNet framework to train expert models for each dataset~(same train/val/test split as VISTA3D) until full convergence. TotalSegmentator is applied out-of-the-box as a foundation model. These three baselines are the well-established ``go-to" options for automatic segmentation, and VISTA3D achieved comparable ``auto" performances and much better performance if minimum human input is available. Meanwhile, our model is much faster than Totalsegmentator~(it has 5 model ensemble) as shown in Table.~\ref{tab:runtime}. In Fig.~\ref{fig:correction}, we show an example of using click points to correct automatic results.  In Fig.~\ref{fig:monkey2}, we present a case of automatic segmentation over a \textbf{monkey} scan, showing the generalizability of the VISTA3D model. More examples are in the supplementary.

\begin{table*}[t]
\caption{Average dice score of the test split in each dataset. TotalSegV2 results are biased towards nnUNet and TotalSegmentator~(the ground truth is generated by the pretrained TotalSegmentator model, which uses nnUNet architecture, and the training data may include our test split). The Bone Lesion is a private dataset with 237 CT scans. Detailed results of all classes are in the supplementary.}
\label{t:results}
\resizebox{1\textwidth}{!}{%
\begin{tabular}{l|rrrrrr}
\toprule

\multicolumn{1}{l}{}    & Auto3dSeg & nnUNet               & TotalSegmentator     & VISTA3D auto & VISTA3D point & VISTA3D auto+point \\
\midrule
MSD03 Hepatic tumor~\cite{antonelli2022medical}    & 0.616                               & 0.617               & -                    & 0.588        & \textbf{0.701}          & 0.687                \\
MSD06 Lung tumor~\cite{antonelli2022medical}       & 0.562                               & 0.554              & -                    & 0.613       & 0.682          & \textbf{0.719}                 \\
MSD07 Pancreatic tumor~\cite{antonelli2022medical} & 0.485                               & 0.488              & -                    & 0.324       & 0.603          & \textbf{0.638}                \\
MSD08 Hepatic tumor~\cite{antonelli2022medical}    & 0.683                               & 0.659            & -                    & 0.682       & 0.733          & \textbf{0.757}                \\
MSD09 Spleen~\cite{antonelli2022medical}           & 0.965                               & \textbf{0.967}             & 0.966                & 0.952       & 0.938          & 0.954                \\
MSD10 Colon tumor~\cite{antonelli2022medical}     & 0.475                               & 0.473                & -                    & 0.439       & 0.609          & \textbf{0.633}               \\
Airway~\cite{stoverud2023aeropath}       & 0.896                               & \textbf{0.899}                & -                    & 0.852       & 0.819           & 0.867                \\
Bone Lesion         & 0.343                               & 0.396                & -                    & 0.491       & 0.536          & \textbf{0.585}                \\
BTCV-Abdomen~\cite{roth9data}            & 0.807                               & 0.825               & 0.846                & 0.849       & 0.815          & \textbf{0.859}                \\
BTCV-Cervix~\cite{roth2015deeporgan}             & 0.598                               & 0.640                & 0.611               & 0.672       & 0.736          & \textbf{0.775}                \\
VerSe~\cite{sekuboyina2021verse}                   & 0.786                               & 0.828                & 0.832                & 0.825       & 0.896          & \textbf{0.906}                \\
AbdomenCT-1K~\cite{ma2021abdomenct}            & 0.934                              & 0.939          & 0.921                & 0.935        & 0.903          & \textbf{0.940}                \\
AMOS22~\cite{ji2022amos}                  & 0.854                              & 0.854            & 0.824                & 0.841       & 0.785          & \textbf{0.856}                \\
TotalSegV2~\cite{TotalSegmentator}              & 0.882                               & *0.906                 & \textbf{*0.942}      & 0.893       & 0.884          & 0.918                \\
\bottomrule
Average & 0.706 & 0.718 & - & 0.711        & 0.760            & \textbf{0.792}       \\
\bottomrule
\end{tabular}
}
\end{table*}

\begin{table}
    \centering
    \caption{Inference speed on a 16GB V100 GPU with varying image sizes (size after resampled to 1.5x1.5x1.5mm). Default 118 class automatic segmentation using VISTA3D and default 117 class segmentation using TotalSegmentator~\cite{TotalSegmentator}. No special model optimization (e.g. tensorRT).}
    \label{tab:runtime}
    \resizebox{0.45\textwidth}{!}{%
    \begin{tabular}{|c|c|c|c|c|}
    \toprule
        Size & 333x333x603 & 512x512x512 & 512x512x768 & 1024x1024x512 \\ \hline % & 1024x1024x768 \\ \hline
        VISTA3D & 1m07s & 2m09s	& 3m25s	& 9m20s \\ \hline %& killed \\ 
        TotalSeg. & 4m34s & 12m01s & 18m56s & 40m13s \\ %& 55m02s\\
    \bottomrule 
    \end{tabular}
    }
\end{table}

\begin{figure*}
    \centering
    \resizebox{1\textwidth}{!}{%
    \begin{tabular}{ccc}
        % \multicolumn{3}{c}{
        %     \includegraphics[width=0.5\textwidth]{F/mouse.png} \hspace{0.05\textwidth}
                 % \includegraphics[width=0.5\textwidth]{F/Mouse_Micro-CT_Thorax_Left_Lung.png}
        % } \\
        \includegraphics[width=0.5\textwidth]{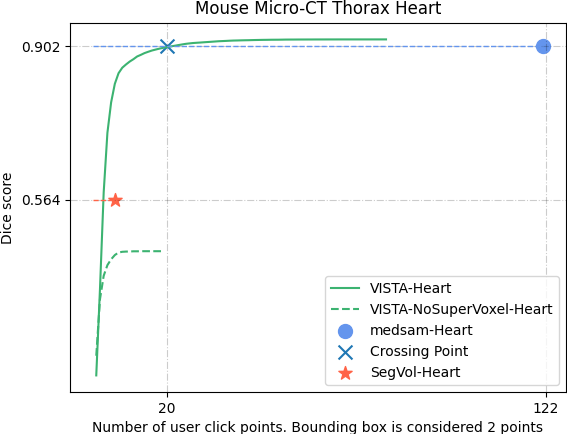} &
         \includegraphics[width=0.5\textwidth]{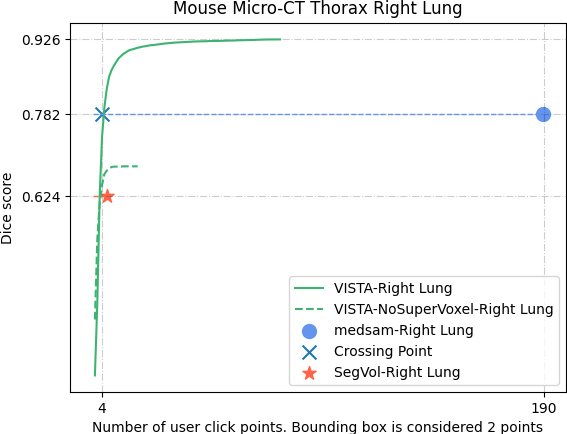} &
         \includegraphics[width=0.5\textwidth]{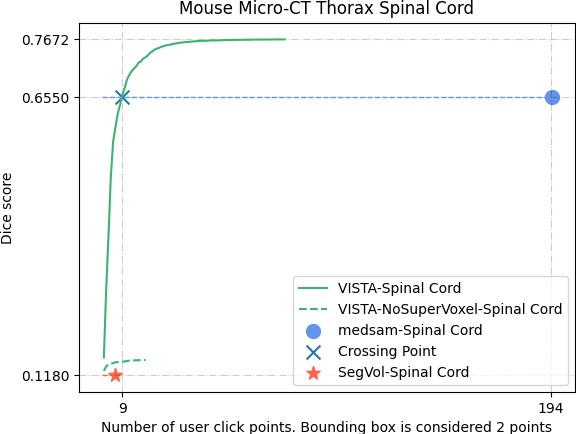} \\
         \includegraphics[width=0.5\textwidth]{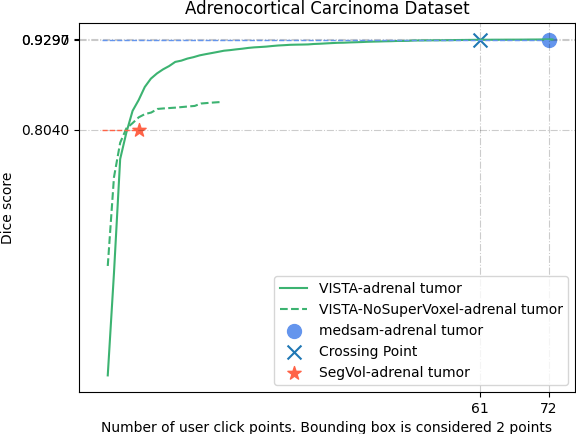} &
         \includegraphics[width=0.5\textwidth]{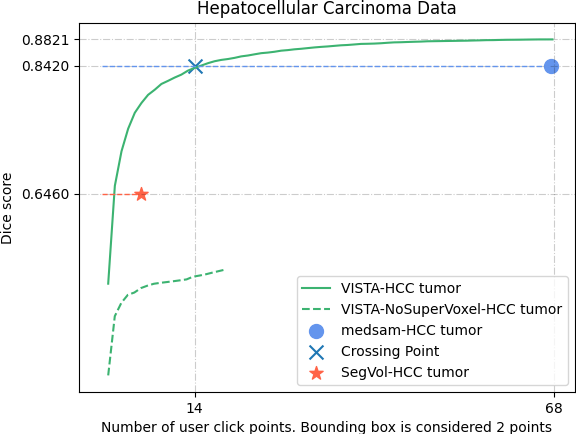}  &
    \includegraphics[width=0.5\textwidth]{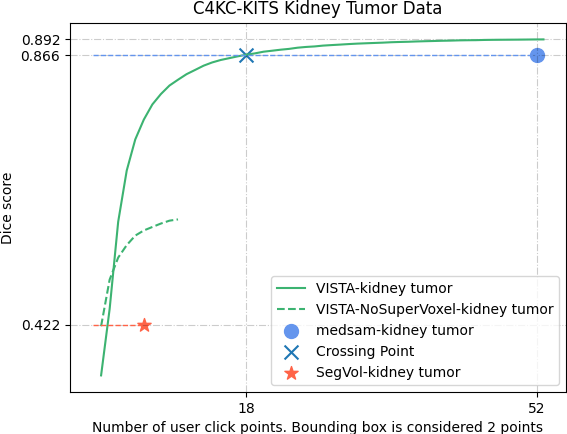}
    \end{tabular}
    }
    \caption{Zero-shot dice scores. The X-axis is the number of click points. The Y-axis is the average dice score over the whole dataset.}
    \label{fig:zeroshot}
\end{figure*}
\begin{figure*}[!hbt]
    \centering
    \includegraphics[width=1\textwidth]{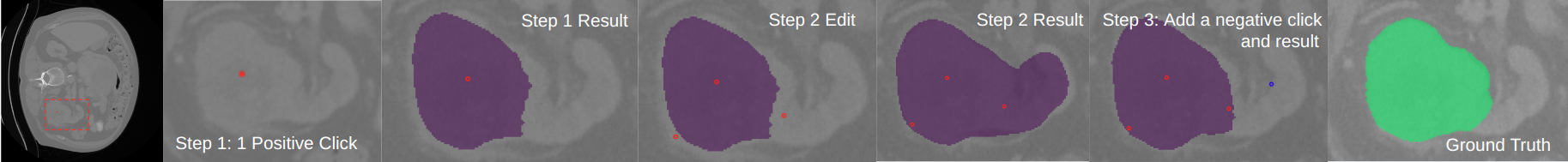}
    \caption{The fine-grained zero-shot interactive segmentation on kidney tumor. The first figure shows the region of the tumor. Step 1 click a positive point~(red) on the tumor and get the results. Step 2 click more points to refine the details. The result has over-segmentation and add a negative point~(blue) on step 3 to get the final results.}
    \label{fig:clicks}
\end{figure*}

\begin{figure}[!hbt]
    \centering
    \includegraphics[width=0.45\textwidth]{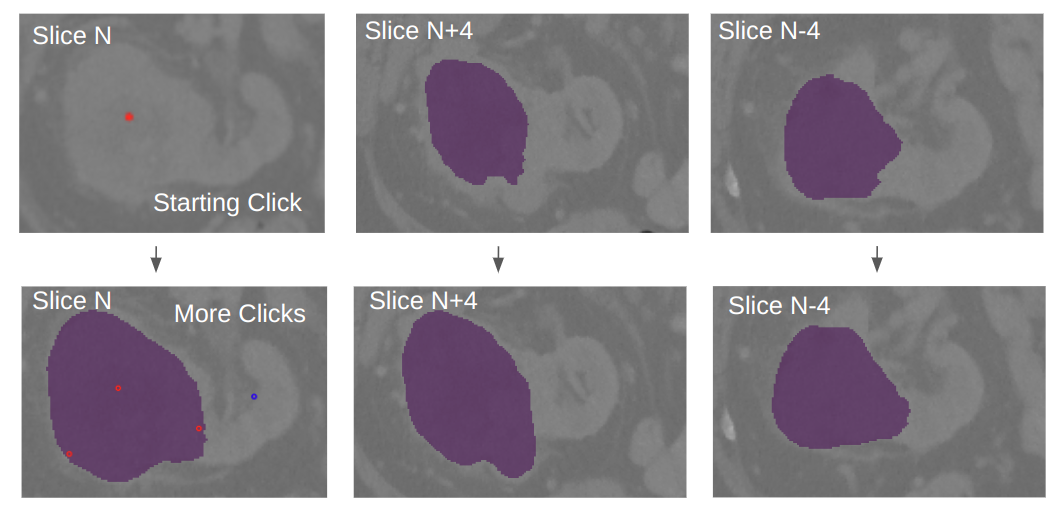}
    \caption{The adjacent slices~(slice N+4 and slice N-4) responses to the clicks on slice N~(the same slice and click on Fig.~\ref{fig:clicks}). The results show how the clicks affect 3D space.}
    \label{fig:3dclicks}
\end{figure}

\section{Zero-shot results}
In this section, we test the zero-shot ability of VISTA3D. We compare with MedSAM~\cite{MedSAM} and SegVol~\cite{du2023segvol} as they showed the best interactive performances in 2D and 3D separately. 
For the MedSAM baseline, we adopt the 3D inference pipeline via a series of 2D slices as described in~\cite{MedSAM}. For segmentation targets that are larger than 10 voxels, tight bounding boxes for each slice were generated to simulate user-provided prompts. Each bounding box is considered the same annotation effort as two-point prompts in our evaluations. For the SegVol baseline, the default settings~\cite{du2023segvol} are evaluated using a positive point with three pairs of positive and negative points~(7 points in total), as well as the zoom-out-zoom-in inference strategy. For VISTA3D,
We mimic user annotations to perform iterative point clicks. The first point is sampled at the foreground center, then the next point will be randomly sampled from the largest connected false positive or false negative region, which has a larger area size.  We evaluate the performance of 4 external datasets with novel classes that our automatic segmentation does not support. 1) the murine dataset~\cite{malimban2022deep} includes 140 micro-CT scans (0.2mm resolution) with 4 annotated mouse organs: heart, left lung, right lung, and spinal cord. The left lung showed similar results to the right lung, and the example mouse CT scan and left lung results are shown in the appendix. 2) the C4KC-KITS~(kidney tumor, 210 scans) dataset~\cite{heller2019data}, the Adrenocortical Carcinoma~(53 scans)  dataset~\cite{moawad2023voxel,clark2013cancer}, the Hepatocellular Carcinoma~(105 scans) dataset~\cite{morshid2019machine,clark2013cancer}. The results are shown in Fig.~\ref{fig:zeroshot}. The results show the superior performance of VISTA3D in both accuracy and reduced annotation efforts. VISTA3D trained without supervoxel~(VISTA-NoSupervoxel) is also shown in the figure, and the results showed the importance of supervoxel for the zero-shot ability.
In Fig.~\ref{fig:clicks}, we show the iterative point clicks of a kidney tumor from C4KC-KITS dataset~\cite{heller2019data}. From step 1 to step 3 we can see how the segmentation responds to the clicks. Fig.~\ref{fig:3dclicks} shows the responses on other slices without clicks. This shows how the clicks respond in 3D space.

\section{Finetuning results}
In many cases, interactive annotation is used to curate enough data to train an accurate automatic model. We show the potential of transfer learning using VISTA3D pretrained checkpoint.
We perform finetuning on the automatic branch under the setting of one-shot, five-shot, and until the full training data split. The dataset we use includes the Whole abdominal Organs Dataset (WORD)~\cite{word2022} and the micro-CT mouse dataset~\cite{malimban2022deep}. We compare with training from scratch methods~(nnU-Net and Auto3DSeg, default setting), and finetuning Totalsegmentator pretrained checkpoint with default nnU-Net pipeline. We use the train/val/test data split from WORD~\cite{word2022} and our own split for the mouse dataset. The results of the held-out test set are shown in Table~\ref{table:finetune}. Compared with the baselines, VISTA3D showed much better performance under few shot setting. Meanwhile, the WORD results of using the full training split~(100 cases) are directly comparable with all the baselines in WORD~\cite{word2022} paper and VISTA3D has the highest dice score~(0.875) over all 10 baselines. The results support our claim where users can annotate a few examples and finetune VISTA3D to build a data flywheel.   
% \usepackage{multirow}
\begin{table}
\caption{Finetuning performances of average test dice scores, with respect to the number of training cases.}
\label{table:finetune}
\centering
\resizebox{0.44\textwidth}{!}{%
\begin{tabular}{l|c|cccc}
\toprule
Dataset                         & \# of cases  & Auto3DSeg & nnU-Net & TotalSegmentator & VISTA3D \\ \hline
\multirow{6}{*}{\shortstack{Micro-CT\\Mouse}} & 1                     & 0.820     & 0.759  & 0.791            &  \textbf{0.926}   \\
                                & 5                     & 0.923     & 0.922  & 0.924            & \textbf{0.935}   \\
                                & 10                    & 0.934     & 0.930  & 0.936            & \textbf{0.938}   \\
                                & 20                    & \textbf{0.947}     & 0.942  & 0.944            & 0.944   \\
                                & 40                    & \textbf{0.949}     & \textbf{0.949}  & \textbf{0.949}            & 0.948   \\
                                & 89                    & 0.949     & 0.949  & \textbf{0.951}            & \textbf{0.951}   \\ \hline
\multirow{6}{*}{WORD}           & 1                     & 0.214     & 0.185  & 0.779            & \textbf{0.795}   \\
                                & 5                     & 0.611     & 0.562  & 0.823            & \textbf{0.839}   \\
                                & 10                    & 0.744     & 0.697  & 0.837            & \textbf{0.855}   \\
                                & 20                    & 0.806     & 0.793  & 0.855            & \textbf{0.862}   \\
                                & 40                    & 0.862     & 0.831  & 0.857            & \textbf{0.869}   \\
                                & 100                   & 0.873     & 0.874  & \textbf{0.875}            & \textbf{0.875}  \\ \hline
\end{tabular}
}
\end{table}

%% file: sec/4_discussion.tex
\section{Conclusion}
In this paper, we introduced VISTA3D, the first unified 3D CT foundation segmentation model.  All the components of VISTA3D are designed to fulfill our proposed human-in-the-loop workflow, such that VISTA3D can be used out-of-the-box to save human effort. 
It achieved highly accurate segmentation comparable with specialized expert models for each dataset, state-of-the-art interactive segmentation for both zero-shot and results editing, and strong transfer learning ability. The large-scale training data with diverse types of labels, carefully designed model architecture, and training recipes were vital for building this highly capable model. We also utilize the best practices in 3D medical image analysis~(e.g. sliding-window, patches, 3D convolutions, data synthesis) to improve the results.  For future work, we are working on 1) enlarging the supported class number and modalities, including adding supports for MRI and PET imaging, 2) improving the zero-shot experiences by developing smarter methods to better utilize the datasets and model checkpoints from natural images, and 3) validating and integrating the workflow with clinical partners.

%% file: sec/X_suppl.tex
\clearpage
\setcounter{page}{1}
\maketitlesupplementary

\section{Dataset Details}
\begin{figure}
    \includegraphics[width=0.45\textwidth]{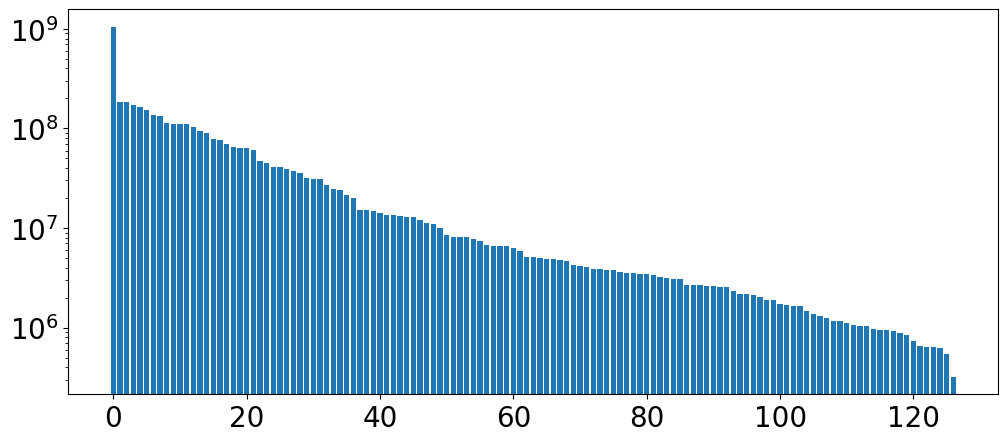} 
    \caption{Distribution of annotated voxels in the training set (X-axis: class index, Y-axis: number of annotated voxels per class).}
    \label{fig:class_hist}
\end{figure}

Table~\ref{table:training_dataset} lists more details about our curated dataset. Fig.~\ref{fig:class_hist} shows the number of annotated voxels according to the corresponding task classes. Spatial resolutions range from $0.45\times0.45\times0.45$ to $1.50\times1.50\times7.50$ (median: $0.88\times0.88\times1.50$) $mm^3$. 

\noindent\textbf{Global and local index for the partial label} Those datasets have different number of classes and indexes in their manual labels~(e.g. Pancreas in MSD07 has index 1 but 10 in TotalsegmentatorV2). We curated a global index of 127 integers and mapped all local indexes in each individual dataset to this global index. We also curated a \textbf{label set} list for each dataset, containing the class index that will be used within this dataset. We included as much dataset with a commercial license as possible for the development of this method.

\begin{table}[H]
\centering
\resizebox{0.47\textwidth}{!}{%
\begin{tabular}{||l l c||} 
 \hline
 Dataset ID & Type & \# cases used \\
 \hline\hline
  TCIA Pancreas CT~\cite{roth9data} & Abdominal CT organs & 80 \\
  AbdomenCT-1K~\cite{ma2021abdomenct} & Abdominal CT organs & 1,050 \\
  AMOS22~\cite{ji2022amos} & Abdominal CT organs & 300 \\
  MSD Task 3,6,7,8,9,10~\cite{antonelli2022medical}  & Various lesions & 945 \\
  CT-ORG~\cite{rister2020ct} & Lung, bones, liver, kidneys, bladder & 136 \\
  TotalSegmentator~\cite{TotalSegmentator} & Many anatomic structures & 1,228 \\
  CRML-CT~\cite{simpson2024preoperative} & Liver, colorectal liver metastases & 197 \\
  % C4KC-KiTS~\cite{heller2019data} & Kidneys, kidney tumor & 210 \\
  VerSe~\cite{sekuboyina2021verse} & Vertebral labelling & 374 \\
  AeroPath~\cite{stoverud2023aeropath} & Airways and lungs & 27 \\
  Bone lesion (in-house) & bones & 296 \\
  LIDC-IDRI~\cite{armato2015data} & Unannotated, lung cancer screening thoracic & 470 \\
  COVID-19~\cite{harmon2020artificial} & Unannotated, chest & 524 \\
  TCIA Colonography~\cite{johnson2008accuracy} & Unannotated, abdomen & 1,440 \\
  StonyBrook COVID19 CT~\cite{saltz2021stony} & Unannotated, chest & 1,274 \\
  NLST~\cite{national2011reduced} & Unannotated, chest & 3,113 \\
 \hline
\end{tabular}
}
\caption{Summary of datasets used for model training.}
\label{table:training_dataset}
\end{table}

\section{Computational Details}
\noindent \textbf{Training Requirements} The model is trained on 64 32GB NVIDIA V100 GPUs with around 20,000 total GPU hours. The prompt number~(object class) in a single training iteration is 36 for automatic branch, and 4 for point branch. The model can be trained with 16GB memory GPUs or even lower by reducing the prompt number in each iteration, at the cost of longer training time if number of classes is large.

\begin{figure}
    \centering
\includegraphics[width=0.5\textwidth]{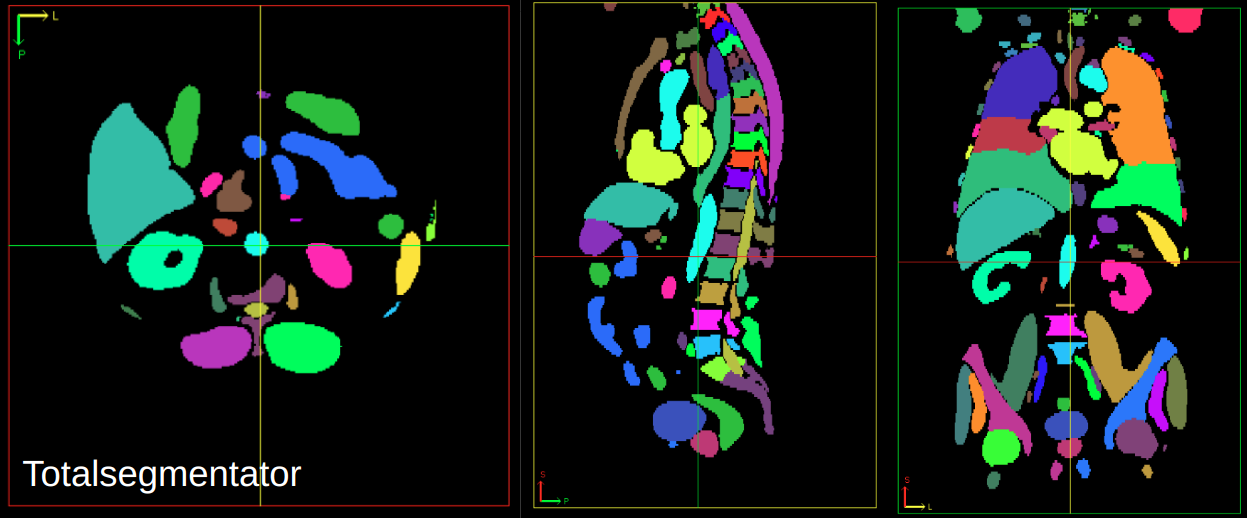}
\includegraphics[width=0.5\textwidth]{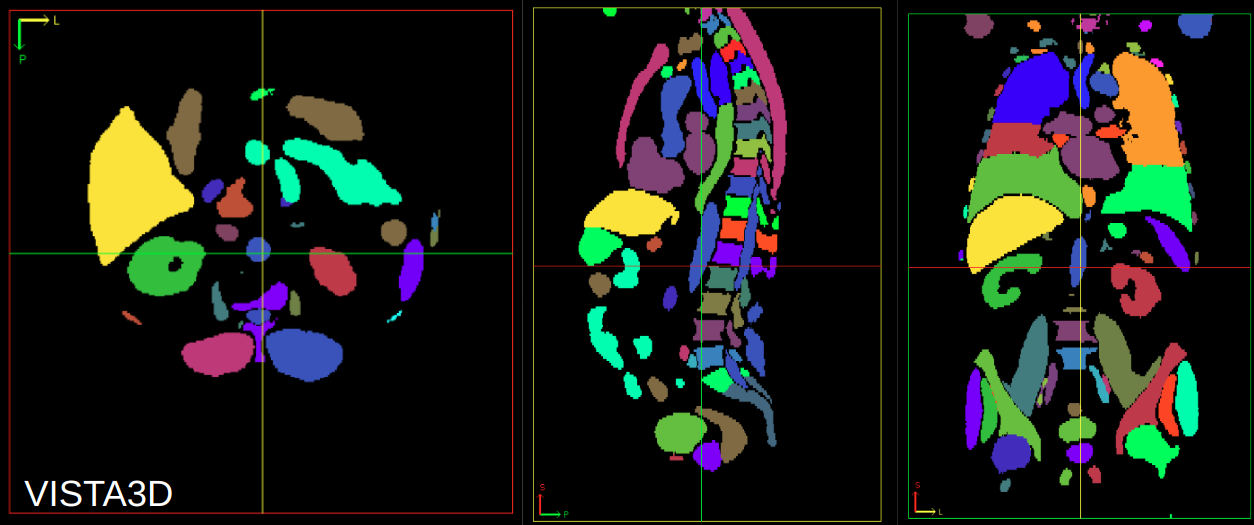}
    \caption{An example of whole class segmentation on a typical sized human CT scan. Running on a lower end machine with 12GB GPU. The runtime for VISTA3D is 1m43s and 2m41s for Totalsegmentator.}
    \label{fig:seg-all}
\end{figure}

\begin{figure}
    \centering
\includegraphics[width=0.5\textwidth]{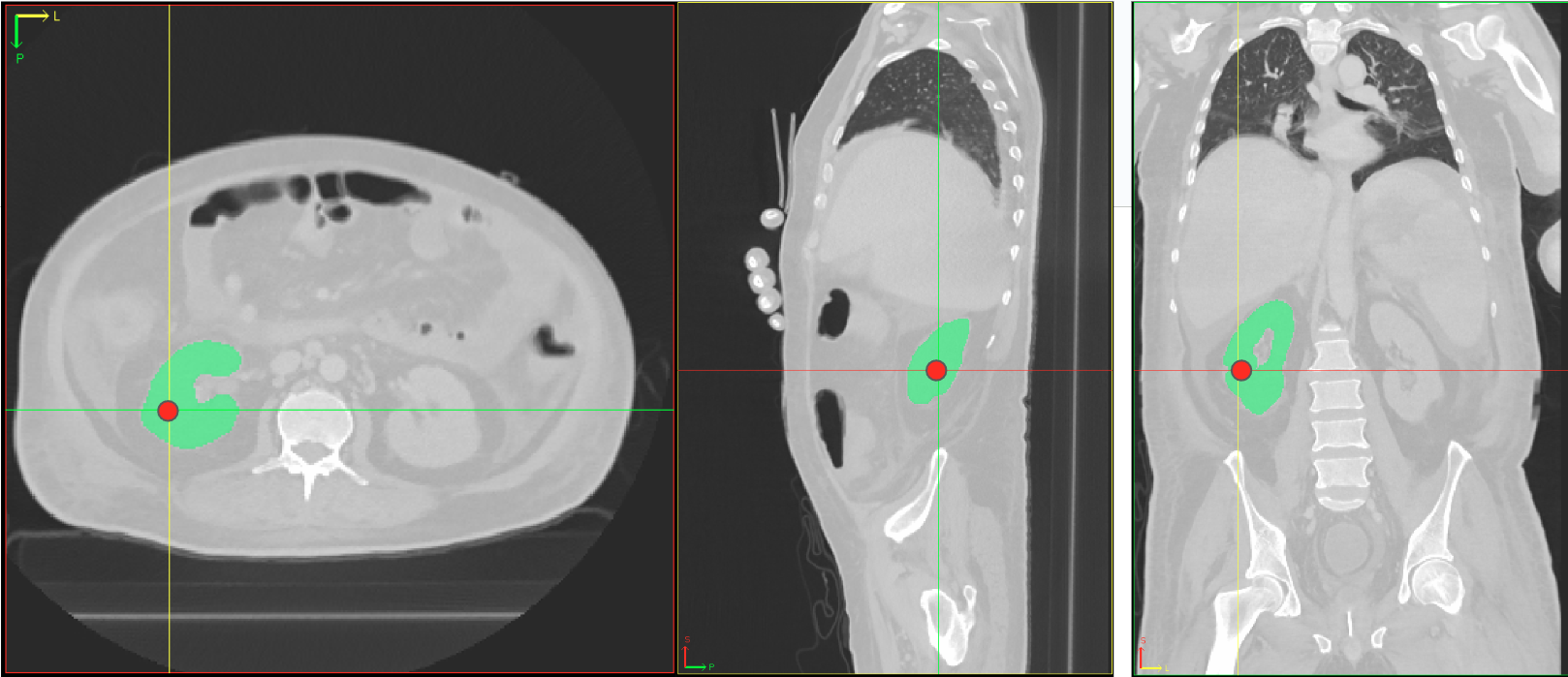}
\includegraphics[width=0.5\textwidth]{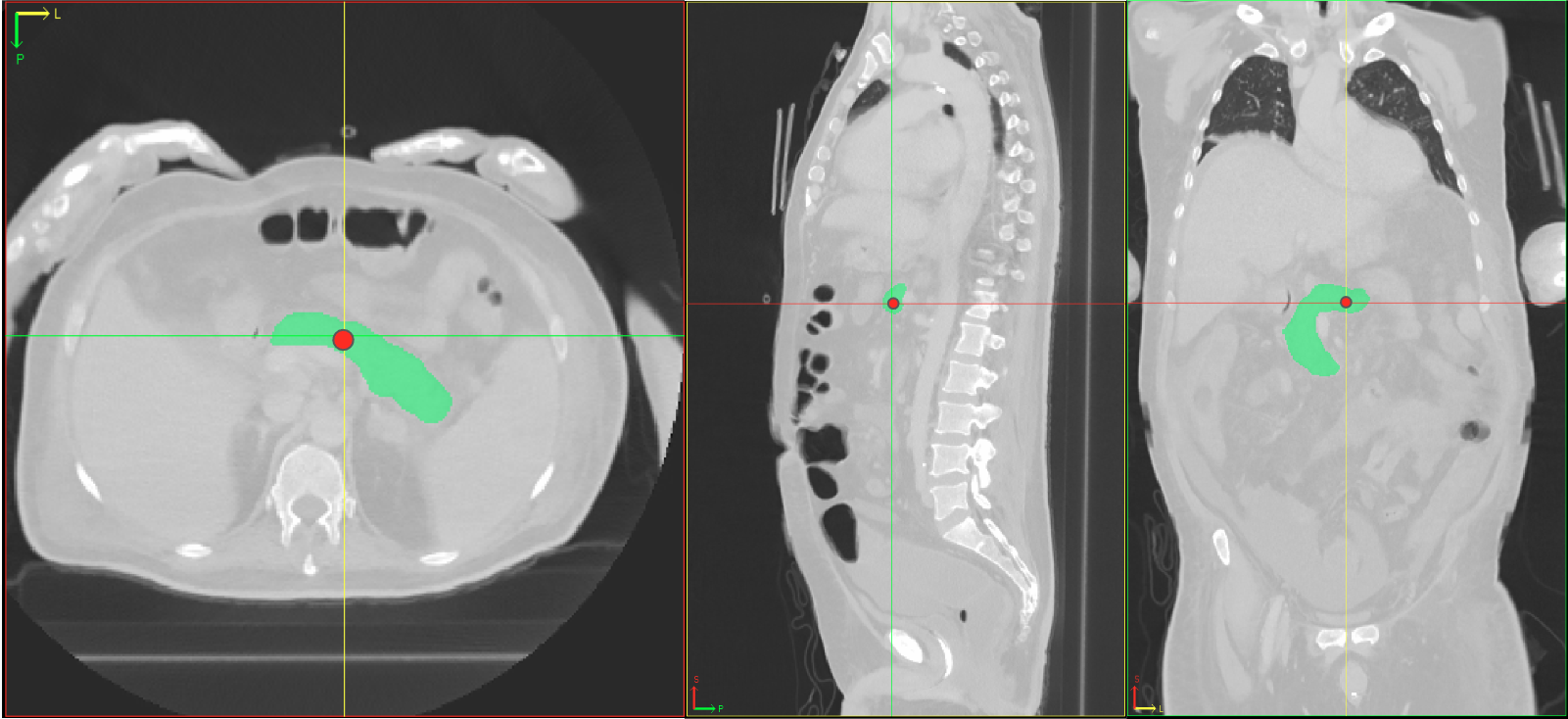}
    \caption{An example of using single point click for organ segmentation. Runtime on a machine with 12GB GPU and 32GB CPU is constant 3.2 seconds, regardless of image size or organ size.}
    \label{fig:click}
\end{figure}

\noindent \textbf{Inference Requirements} The inference GPU memory requirements also depend on the prompt number and image size. Since the model is based on sliding window of size 128x128x128, the GPU memory requirements can be optimized to be stable and less dependent on image size. We used a sliding window inferer with adaptive memory control to switch between CPU and GPU to avoid the out-of-memory issue.
We benchmarked the runtime on a 16GB V100 GPU in the main paper. Totalsegmentator uses 5 sub-task models for different class groups and thus can be slower. Here we also performed inference on a lower-end environment with 12GB memory GPU and 32GB memory CPU. The results of a typical CT scan (a MSD task03 test scan~\cite{antonelli2022medical}, size 308x260x453 after resampling) are shown in Fig.~\ref{fig:seg-all}. The runtime for VISTA3D is 1m43s and 2m41s for the Totalsegmentator. For interactive segmentation, the single-click point inference run-time is 3.2s on the same 12GB GPU machine. Two examples are shown in Fig.~\ref{fig:click}.

% \begin{table*}[]
%     \centering
%     \begin{tabular}{|c|c|c|c|c|}
%     \toprule
%         Volume size at 1.5x1.5x1.5 mm & 333x333x603 & 512x512x512 & 512x512x768 & 1024x1024x512 \\ \hline % & 1024x1024x768 \\ \hline
%         VISTA3D RunTime & 1m07s & 2m09s	& 3m25s	& 9m20s \\ \hline %& killed \\ 
%         TotalSegmentator RunTime & 4m34s & 12m01s & 18m56s & 40m13s \\ %& 55m02s\\
%     \bottomrule 
%     \end{tabular}
%     \caption{Benchmarks on a 16GB V100 GPU with varying image size, performing default automatic segmentation of 118 classes of VISTA3D and 117 classes segmentation using TotalSegmentator. Default model inference without optimization (e.g. tensorRT).}
%     \label{tab:runtime}
% \end{table*}
\section{Additional training details}
\subsection{Stage1-Interactive branch training}
The algorithm is shown in Alg.~\ref{alg:train}. The point sampler $S$ works as a data augmenter, with 50\% probability to sample points directly from $y$ to get the point $p$ and binary groundtruth mask $y_{gt}$ as a training pair, while another 50\% will be used with the following augmentations: a) random sample points from supervoxel and form a zero-shot training pair. b) random add or subtract a supervoxel mask that satisfies a certain size and position criterion to $y$, this is used to force the model to be able to edit supported class mask. Meanwhile, when the subtraction or addition size exceeds a certain limit, the generated training pair will also be used as zero-shot pairs with the zero-shot embedding. We use $max_{iter}=5$ for the training.
\begin{algorithm}
\caption{Interactive branch training}\label{alg:train}
\begin{algorithmic}
\Require VISTA interactive branch model $\Phi$, image patch $x$, image manual label $y$, image pseudo label $y_p$, supervoxel $y_s$. 
\Ensure At least one of $y$ or $y_p$ are not None
\State $S \gets point\_sampler(y, y_s) $ \Comment{Initialize point sampler based on manual label and supervoxel}
\State $S_p \gets point\_sampler(y_p, y_s) $ 
\State $ p, y_{gt} \gets S.sample() $ \Comment{Sample point prompts $p$ and segmentation mask $y_{gt}$}
\State $ p^p, y^p_{gt} \gets S.sample() $ 
\For{$i=1$ to $max\_iter$}
\State $loss \gets LossFunction(\Phi(x, p), y_{gt})$ 
\State $loss_p \gets LossFunction(\Phi(x, p^p), y^p_{gt})$ 
\State update $\Phi$ using $loss + loss_p$
\State $p = p \cup Sample(\Phi(x, p), y_{gt})$ \Comment{Sample 1 point each from false positive and negative region}
\State $p^p = p^p \cup Sample(\Phi(x, p^p), y^p_{gt})$
\EndFor
\end{algorithmic}
\end{algorithm}
\subsection{Stage3-Automatic branch training}
For each patch, we randomly sample the existing class indexes $c$ from its manual label or pseudo-label and obtain the corresponding binary mask $y_{gt}$ or $y^p_{gt}$. The algorithm is shown in Alg.~\ref{alg:train_auto}.  Unlike traditional segmentation models that do softmax on multichannel output, our automatic segmentation is based on promptable binary segmentation, thus prone to produce false positives. We mitigate this issue by sampling the background prompts from $label\_set - y.unique()$ or $label\_set - y^p.unique()$ and train the model to produce zero output when responding to the prompt. So in each iteration, a 128 cubic image patch is the model input, and we sample a maximum of 32 class prompts using Alg.~\ref{alg:train_auto} and a maximum of four background prompts. All of those prompts are concatenated in the batch dimension.
\begin{figure}
    \centering
    \includegraphics[width=0.5\textwidth]{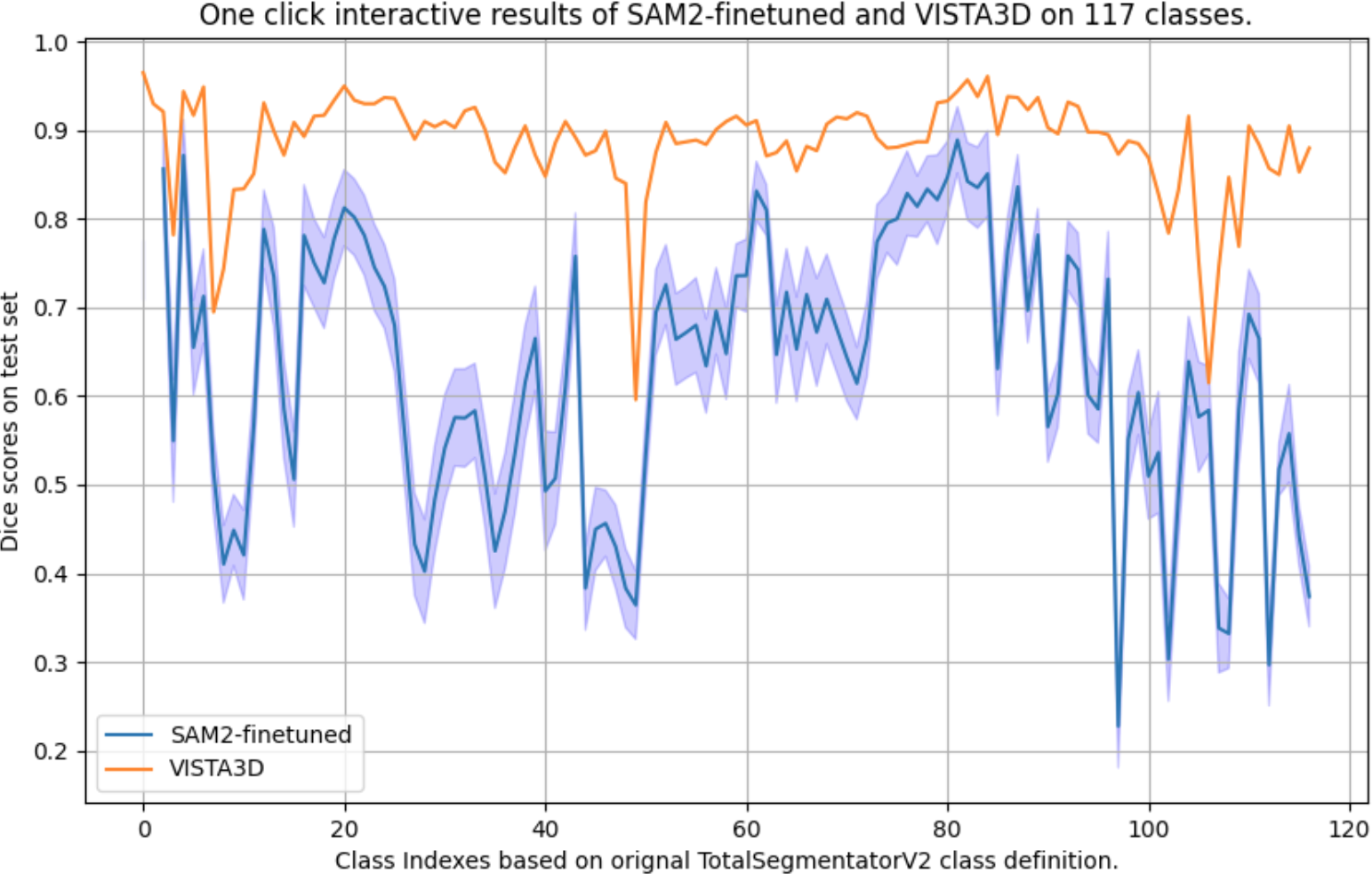}
    \caption{One-point interactive results for SAM2 and VISTA3D on TotalSegmentatorV2 test split. SAM2 is finetuned on the training split with SAM2's official finetuning script. Click point selected at the center slice of each foreground.}
    \label{fig:SAM2_finetuned}
\end{figure}
\subsection{SAM2 finetuning experiments}
\begin{figure}
    \centering
    \includegraphics[width=0.4\textwidth]{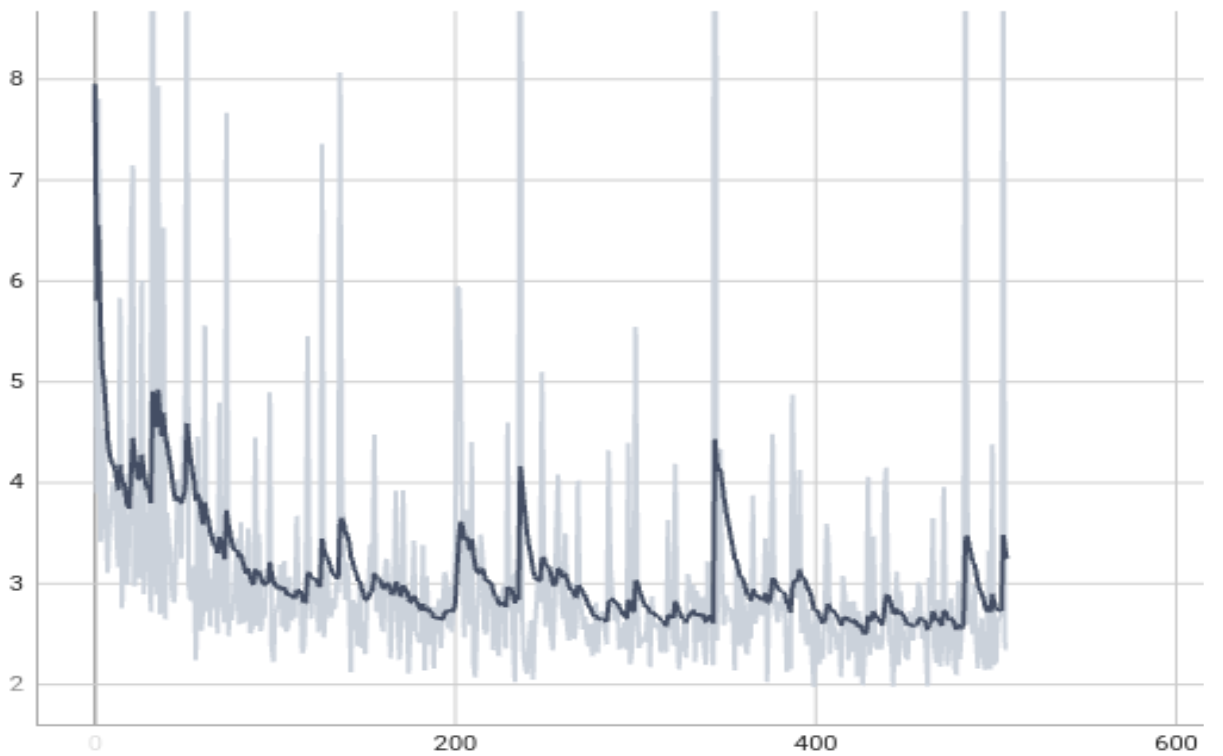}
    \caption{Loss curves for finetuning SAM2 on TotalSegmentator dataset.}
    \label{fig:samfinetune}
\end{figure}

We performed a detailed fine-tuning experiment to see if the SAM2 object tracking-based algorithm can be applied to 3D medical image segmentation. We used the official SAM2 finetuning code\footnote{\url{https://github.com/facebookresearch/sam2/tree/main/training}} and finetuned on the TotalSegmentator~\cite{TotalSegmentator} training set. Each axial slice is considered a video frame. The model is trained on 8 80GB A100 GPU for 500 epochs until full convergence as shown in Fig.~\ref{fig:samfinetune}. However, the results as shown in Fig.~\ref{fig:SAM2_finetuned} is disappointing. We also finetuned on MSD Task07 pancreas and pancreas tumor to reduce the class number. The results can be shown in Table.~\ref{tab:task07}. The SAM2 method can track objects with simple shape and clear boundary very well, like femur bones, but failed to track complicated 3D shapes. Similar findings can be found in~\cite{he2024short}.
\begin{table}[]
    \centering
    \begin{tabular}{|c|c|c|}
    \toprule
       MSD Task07& Pancreas & Pancreas Tumor \\ \hline
       VISTA3D  & 0.802  & 0.603 \\ \hline
       SAM2-Finetuned  & 0.557 & 0.308 \\
       \bottomrule
    \end{tabular}
    \caption{Single click performance on MSD Task07 test set. SAM2 finetuned only on the training split.}
    \label{tab:task07}
\end{table}
\begin{algorithm}[!hbt]
\caption{Automatic branch training}\label{alg:train_auto}
\begin{algorithmic}
\Require VISTA automatic branch model $\Phi_a$ with encoder frozen, image patch $x$, image manual label $y$, image pseudo label $y_p$.
\Ensure At least one of $y$ or $y_p$ are not None
\State $ c, y_{gt} \gets y.unique().sample() $ \Comment{Sample class prompts $c$ and segmentation mask $y_{gt}$}
\State $ c^p, y^p_{gt} \gets y_p.unique().sample() $ 
\State $loss \gets LossFunction(\Phi_a(x, c), y_{gt})$ 
\State $loss_p \gets LossFunction(\Phi_a(x, c^p), y^p_{gt})$ 
\State update $\Phi_a$ using $loss + loss_p$
\end{algorithmic}
\end{algorithm}
\section{Additional Results}
We provide additional VISTA3D results in this section. The baseline MedSAM~\cite{MedSAM} and SegVol~\cite{du2023segvol} results are from their provided user interface and online hugging-face demo. 
\subsection{Qualitative Results}

\begin{figure*}
    \centering
    \includegraphics[width=1\textwidth]{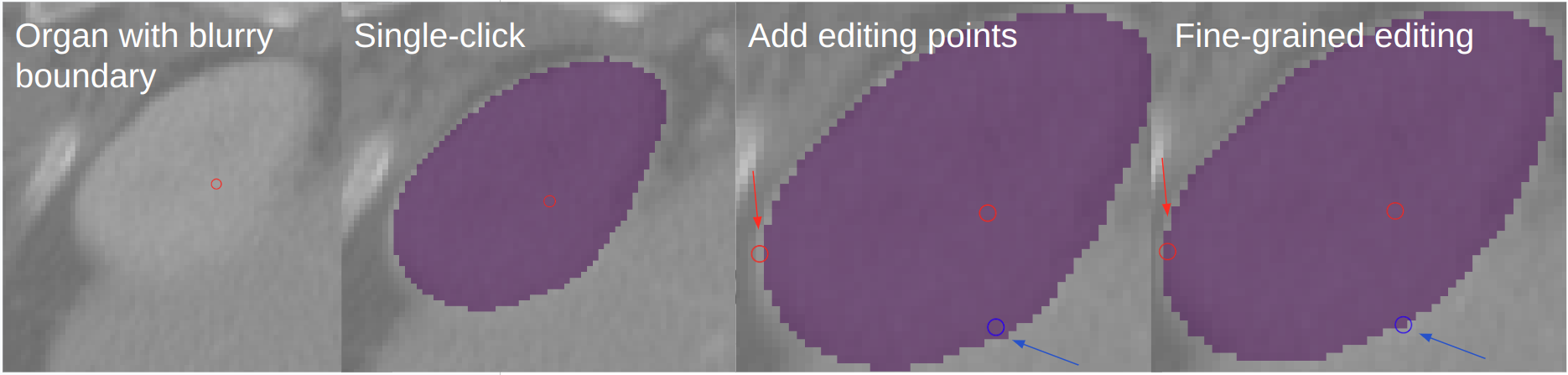}
    \caption{Fine-grained editing on blurry boundary. Red is positive point and blue is the negative point. This is an extreme example to show that VISTA3D can edit one-pixel wide boundaries. The addition or removal area depends on the model's understanding of boundaries, and the edited area by a single click could be much larger.}
    \label{fig:fineedit}
\end{figure*}

\noindent\textbf{Editing examples} We show an extreme example in Fig.~\ref{fig:fineedit}, illustrating that VISTA3D supports detailed editing at pixel level, while the bounding box prompt cannot perform any editing.

\begin{figure*}
    \centering \includegraphics[width=1\textwidth]{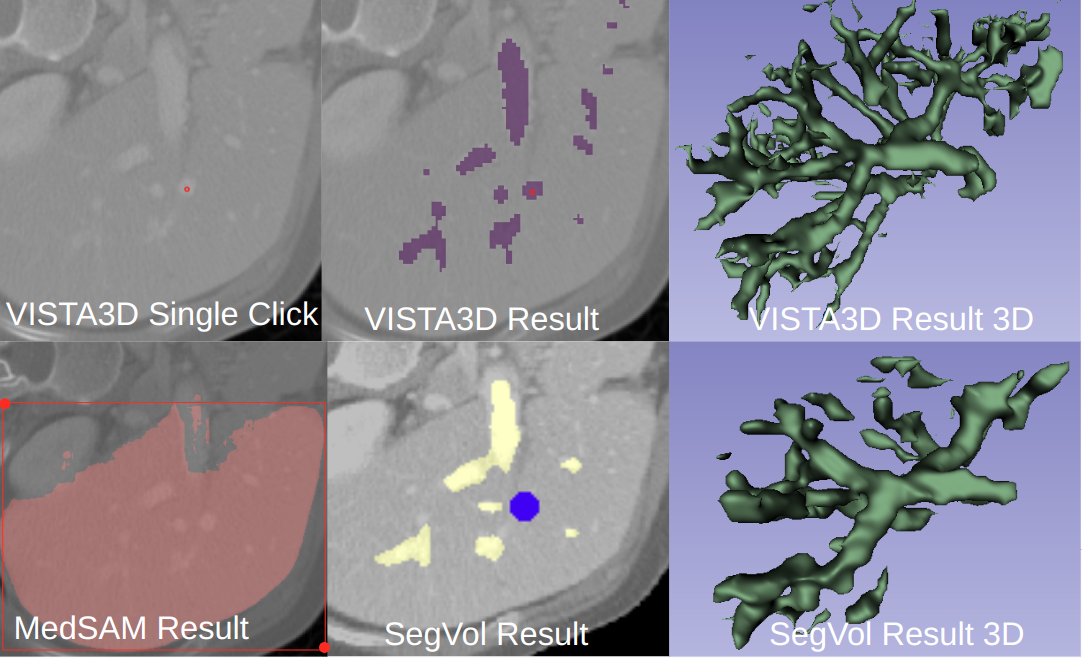}
    \caption{Single positive point for hepatic vessel segmentation (example from MSD09 spleen held out test set, no hepatic vessel groundtruth to avoid groundtruth leakage). SegVol demo uses blue dot while VISTA3D demo uses red dot to represent positive clicks. 
 VISTA3D achieved much better results in details.}
    \label{fig:vessel}
\end{figure*}

\begin{figure*}
    \centering \includegraphics[width=0.85\textwidth]{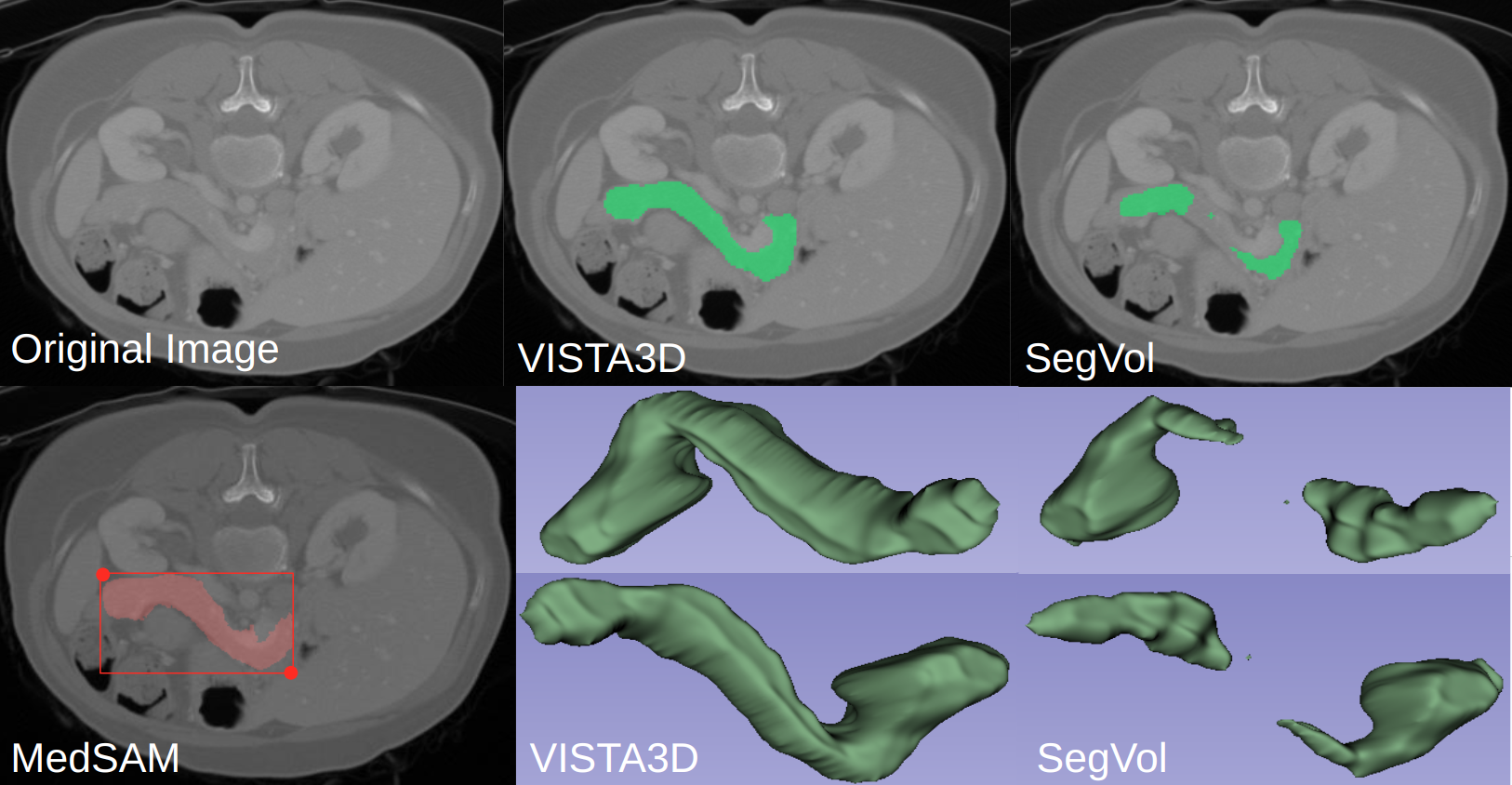}
    \caption{Automatic ~(semantic) segmentation for pancreas (example from MSD09 spleen held out test set, no pancreas groundtruth). VISTA3D achives much better results in details and segmentation completeness.}
    \label{fig:pancreas}
\end{figure*}

\noindent\textbf{Hard Examples} We show some hard classes like hepatic vessel and pancreas in Fig.~\ref{fig:vessel} and Fig.~\ref{fig:pancreas}. Those classes are included in VISTA3D's and SegVol's training sets. We randomly picked an abdominal scan from the MSD task09 spleen test split. This dataset does not contain annotations for pancreas or hepatic vessel, thus it can avoid groundtruth leakage and provide fair comparison.

\begin{figure*}
    \centering
    \includegraphics[width=0.85\textwidth]{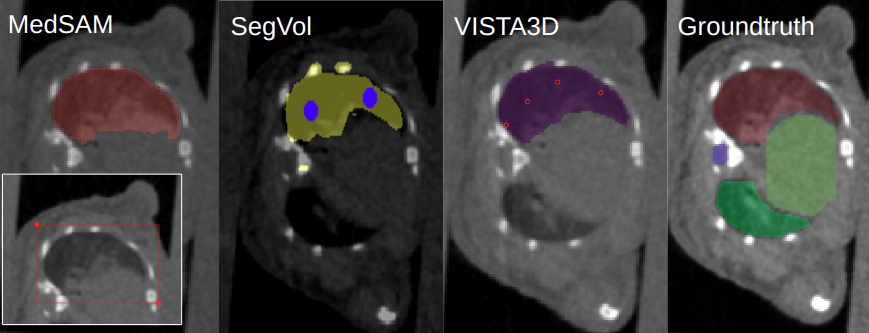}
    \caption{Interactive segmentation on micro-CT mouse left lung. Baseline results from MedSAM local user interface and SegVol demo.}
    \label{fig:medsam_compare}
\end{figure*}

\begin{figure*}
    \centering
    \includegraphics[width=1\textwidth]{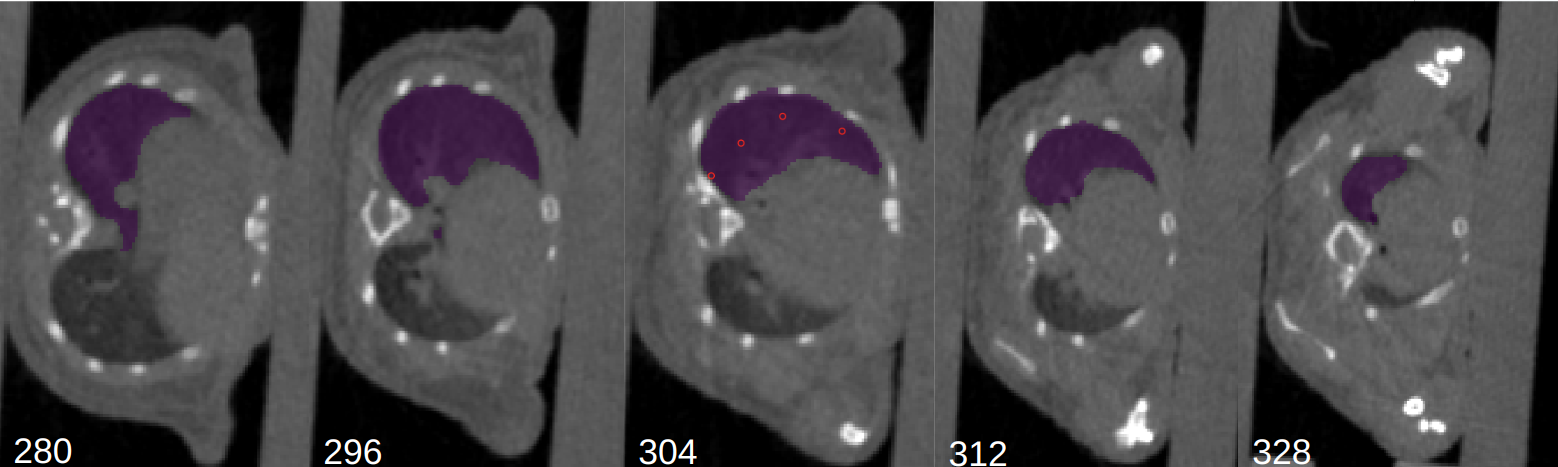}
    \caption{3D point response on far away slices. The point click is on slice 304~(same as Fig.~\ref{fig:medsam_compare}), but the segmentation on slices 280, 296, 312, and 328 all showed good results, showing the potential of reducing annotation effort in 3D space.}
    \label{fig:mouse_slilce}
\end{figure*}

\begin{figure*}
    \centering
    \begin{tabular}{cc}
         \includegraphics[width=0.4\textwidth]{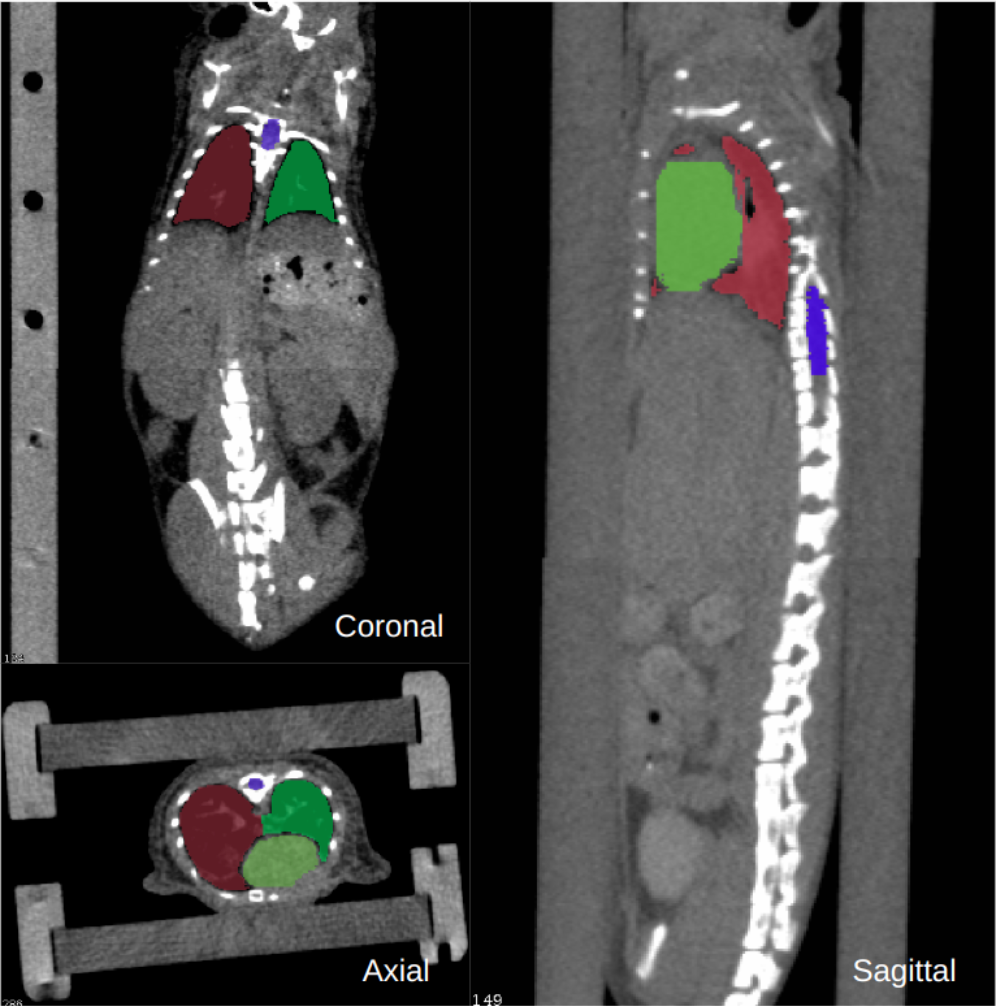} &     \includegraphics[width=0.55\textwidth]{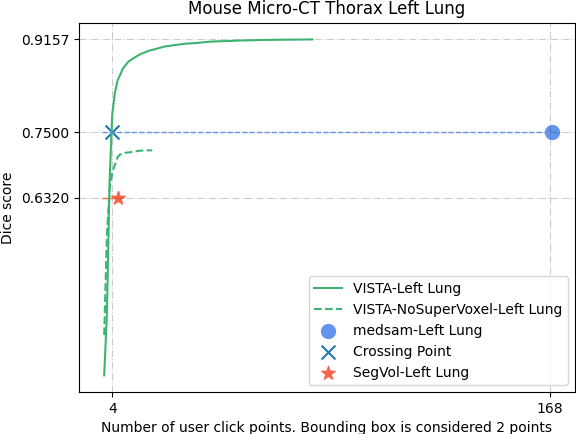}
    \end{tabular}
    \caption{The mouse micro-CT example and the mouse left lung zero-shot performances. Even "left lung" is in the supported class, the huge structural difference between human and mice will fail any automatic segmentation model trained on human anatomy.}
    \label{fig:mouse-sp}
\end{figure*}

\noindent\textbf{Zero-shot interactive examples}  In Fig.~\ref{fig:medsam_compare}, we show the interactive segmentation on a micro-CT mouse left lung. We can see that MedSAM has a major weakness of not being able to perform fine detailed editing, while SegVol's response resolution is low. Fig.~\ref{fig:mouse_slilce} shows other slices of the same mouse scan as Fig.~\ref{fig:medsam_compare}. The figure shows a good point response on slices even far away from the clicks, illustrating the ability of 3D annotation and reducing annotation effort. We also provide additional illustrations of Mouse-CT dataset and our zero-shot results for left lung in Fig.~\ref{fig:mouse-sp}.

\begin{figure*}
    \centering
    \includegraphics[width=1\textwidth]{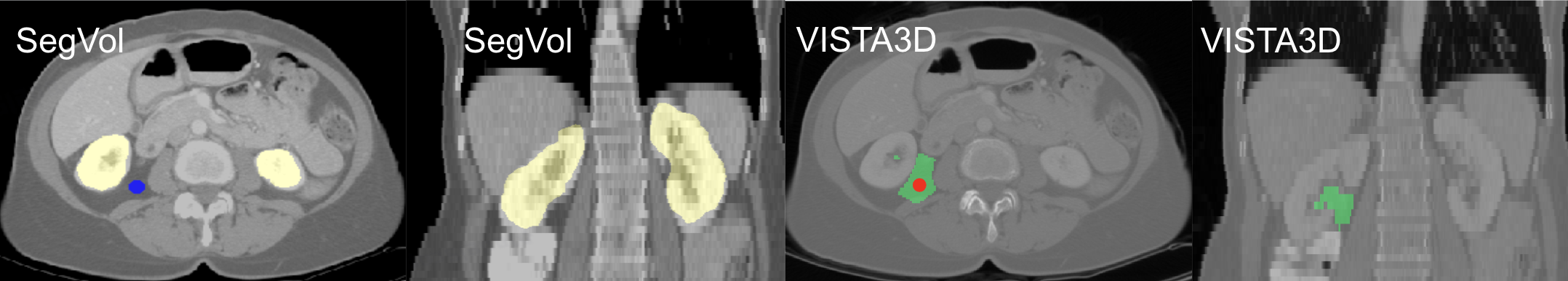}
    \caption{The overfitting problem with common organs. Due to the lack of diversity of 3D organs, the model can easily overfit to certain classes and ignore point clicks, even without providing any semantic information about the class. We click a positive point outside of kidney to segment the fluid around, and this should be zeroshot. SegVol directly overfits to segment kidney and ignores the points. VISTA3D avoided this problem by using the zero-shot embedding and the novel model and recipe design.}
    \label{fig:zeroshot_kidney}
\end{figure*}

\begin{figure*}
    \centering
    \includegraphics[width=1\textwidth]{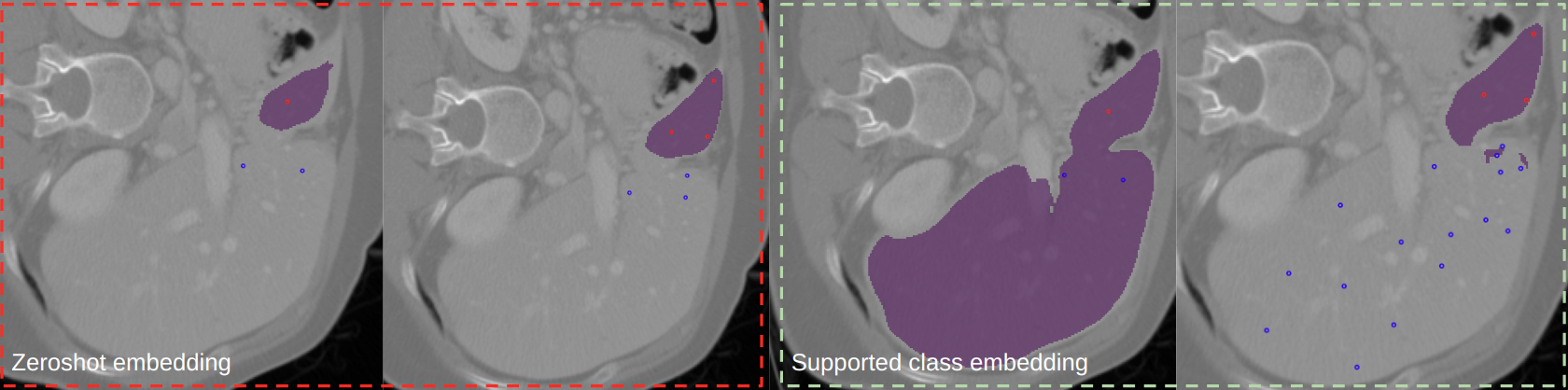}
    \caption{Use points to forcefully separate liver into substructures. We can see that VISTA3D with zero-shot embedding responds much better to the clicks. However, if the model uses supported class embedding, the model is reluctant to respond to negative points for liver segmentation.}
    \label{fig:zeroshotembed}
\end{figure*}

\noindent\textbf{Overfitting to common organs} Due to the lack of diversity of 3D organs, the model can easily overfit to certain classes and remember the shapes, intensities, or locations. This is beneficial for achieving superior segmentation accuracy, but on the other hand, the model will  ignore point clicks, even without providing any semantic information about the class. An example is shown in Fig.~\ref{fig:zeroshot_kidney}. We click a point outside of the kidney to segment the fluid around, and this should be zeroshot. SegVol directly segments the kidney and ignores the point. VISTA3D avoids this problem by using the zero-shot embedding and the novel model and recipe design. The area outside of organ is relatively easy; what if we want to forcefully segment a supported organ into sub-parts? We show an example in Fig.~\ref{fig:zeroshotembed}. If we click positive points on the liver, the model tends to ignore the points and directly segment the liver. Adding a zero-shot embedding will make the model follow the clicks much better.

\subsection{Quantitative Results}
We provide detailed Dice scores on all the classes of our test datasets.
The result is shown in Table.~\ref{t:dice}.

\section{Additional Discussions}
The VISTA3D model design will naturally raise two questions, why not share decoder and why share encoder. If we share the encoder and decoder, then automatic and interactive will be trained together, which will 1) slow down the training. Interactive branch is much more memory intensive than automatic branch, and the supervoxels are only used for interactive training, thus, automatic branch can use a much larger batch size. Combine these two training will reduce automatic branch training iteration and its performance. 2) There are internal conflicts between zero-shot and automatic segmentation, our pilot study showed worse results and our auto-branch is not able to reach state-of-the-art results once trained together with interactive branch. Sharing encoder has two purposes, 1) we support interactive editing over automatic results, the shared encoder could reduce inference computation cost. 2) The interactive branch can be trained with a much broader range of data, thus the encoder can extract more generalizable features and help with the generalizability of automatic segmentation.

\clearpage
\onecolumn
\footnotesize
\begin{longtable}{lllllll}
\caption{Dice score of all the classes on the test datasets.}\\
\label{t:dice}
\tiny

                             & \begin{tabular}[c]{@{}c@{}}\textbf{Auto-} \\ \textbf{3dSeg}\end{tabular} & \textbf{nnUNet} & \begin{tabular}[c]{@{}c@{}}\textbf{TotalSeg-} \\ \textbf{mentator}\end{tabular} & \begin{tabular}[c]{@{}c@{}}\textbf{VISTA3D} \\ \textbf{auto}\end{tabular} & \begin{tabular}[c]{@{}c@{}}\textbf{VISTA3D} \\ \textbf{point}\end{tabular} & \begin{tabular}[c]{@{}c@{}}\textbf{VISTA3D}\\  \textbf{auto + point}\end{tabular} \\
\hline
\multicolumn{7}{l}{\textbf{MSD03 Hepatic Tumor}}                                                                                                                                                                                                                     \\  
liver                        & 0.943     & 0.947  & 0.942            & 0.959                                                   & 0.874                                                    & 0.961                                                           \\
hepatic tumor                & 0.616     & 0.617  & -                & 0.588                                                   & 0.701                                                    & 0.687                                                           \\\hline
\multicolumn{7}{l}{\textbf{MSD06 Lung Tumor}}                                                                                                                                                                                                                        \\
lung tumor                   & 0.562     & 0.554  & -                & 0.614                                                   & 0.682                                                    & 0.719                                                           \\
\hline                                                 
\end{longtable}
\clearpage
\twocolumn